\documentclass{article}

\usepackage[accepted]{icml2023}

\usepackage{hyperref}

\usepackage{microtype}
\usepackage{graphicx}
\usepackage{subfigure}
\usepackage{booktabs} 

\usepackage{amsmath}
\usepackage{amssymb}
\usepackage{mathtools}
\usepackage{amsthm}

\usepackage[capitalize,noabbrev]{cleveref}
\newcommand{\eg}{e.g., }
\newcommand{\ie}{i.e., }

\newcommand{\cf}{cf.\ }

\theoremstyle{plain}
\newtheorem{theorem}{Theorem}[section]

\theoremstyle{definition}
\newtheorem{definition}[theorem]{Definition}

\theoremstyle{remark}

\usepackage{svg}

\icmltitlerunning{Memory-Based Meta-Learning on Non-Stationary Distributions}

\def\cdbar{|}

\newcommand{\cX}{\mathcal{X}}

\newcommand{\cM}{\mathcal{M}}

\newcommand{\cC}{\mathcal{C}}

\newcommand{\cP}{\mathcal{P}}

\newcommand{\cT}{\mathcal{T}}

\newcommand{\ptw}{\text{\sc ptw}}
\newcommand{\lin}{\text{\sc lin}}

\newcommand{\kt}{\text{\sc kt}}

\begin{document}

    \twocolumn[
    \icmltitle{Memory-Based Meta-Learning on Non-Stationary Distributions}



    \icmlsetsymbol{equal}{*}
    
    \begin{icmlauthorlist}
    \icmlauthor{Tim Genewein}{equal,dm}
    \icmlauthor{Gr\'egoire Del\'etang}{equal,dm}
    \icmlauthor{Anian Ruoss}{equal,dm}
    \icmlauthor{Li Kevin Wenliang}{dm}
    \icmlauthor{Elliot Catt}{dm}
    \icmlauthor{Vincent Dutordoir}{dm,ucamb}
    \icmlauthor{Jordi Grau-Moya}{dm}
    \icmlauthor{Laurent Orseau}{dm}
    \icmlauthor{Marcus Hutter}{dm}
    \icmlauthor{Joel Veness}{dm}
    \end{icmlauthorlist}
    
    \icmlaffiliation{dm}{DeepMind}
    \icmlaffiliation{ucamb}{University of Cambridge}
    
    \icmlcorrespondingauthor{Tim Genewein}{timgen@deepmind.com}
    \icmlcorrespondingauthor{Gr\'egoire Del\'etang}{gdelt@deepmind.com}
    \icmlcorrespondingauthor{Anian Ruoss}{anianr@deepmind.com}
    
    \icmlkeywords{Memory-based meta-learning, Bayesian inference, logarithmic loss, sequential prediction}
    
    \vskip 0.3in
    ]
    
    
    
    \printAffiliationsAndNotice{\icmlEqualContribution} 

    \begin{abstract}
    Memory-based meta-learning is a technique for approximating Bayes-optimal predictors.
    Under fairly general conditions, minimizing sequential prediction error, measured by the log loss, leads to implicit meta-learning.
    The goal of this work is to investigate how far this interpretation can be realized by current sequence prediction models and training regimes.
    The focus is on piecewise stationary sources with unobserved switching-points, which arguably capture an important characteristic of natural language and action-observation sequences in partially observable environments.
    We show that various types of memory-based neural models, including Transformers, LSTMs, and RNNs can learn to accurately approximate known Bayes-optimal algorithms and behave as if performing Bayesian inference over the latent switching-points and the latent parameters governing the data distribution within each segment.
\end{abstract}

    \section{Introduction}

Memory-based meta-learning (MBML) has recently risen to prominence due to breakthroughs in sequence modeling and the proliferation of data-rich multi-task domains.
Previous work \citep{ortega2019meta, mikulik2020meta} showed how, in principle, MBML can lead to Bayes-optimal predictors by learning a fixed-parametric model that performs amortized inference via its activations.
This interpretation of MBML can provide theoretical understanding for counter-intuitive phenomena such as in-context learning that emerge in large language models with frozen weights \citep{xie2022explanation}.

\begin{figure*}[!ht]
    \centering
    \includesvg[scale=0.6]{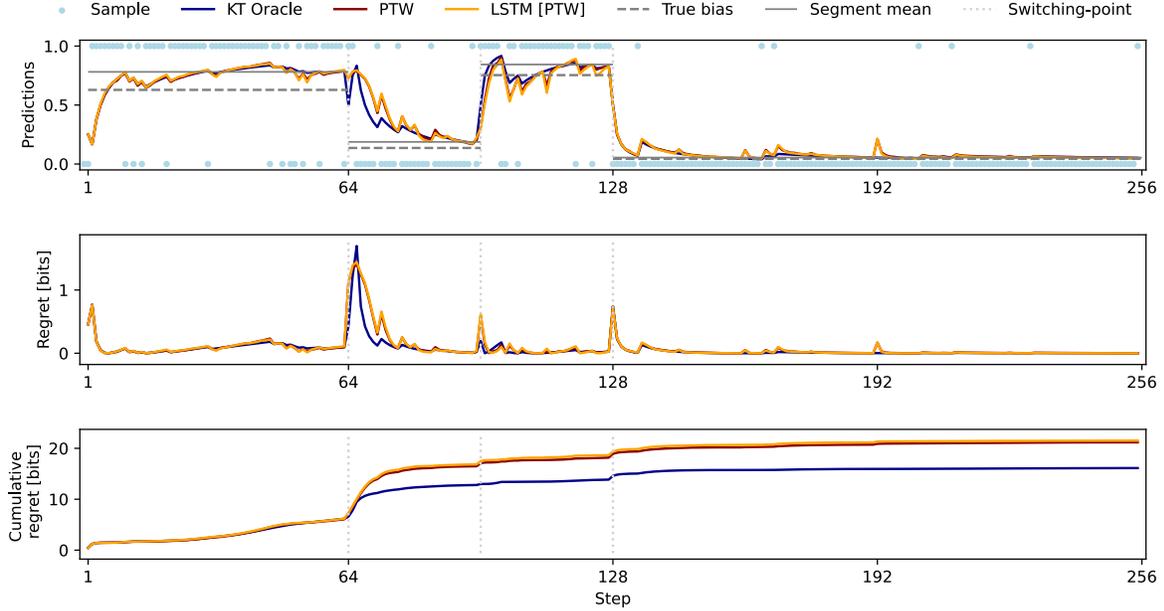}
    \caption{
        A single sequence from a piecewise Bernoulli source with three switching-points drawn from the \ptw{} prior (see \cref{sec:methods}).
        Top: The predictors observe streams of binary samples~$x_t$ and, at each step, predict the probability of the next observation.
        The solid lines show predictions~$p(x_t\vert x_{<t})$ by the Bayes-optimal \ptw, the KT Oracle that observes switching-points, and the trained LSTM (trained on data from \ptw{} prior, indicated in the square bracket).
        Both the LSTM and \ptw{} rapidly adapt after switching-points, enabled via the inductive bias of the \ptw{} prior and acquired by the LSTM via meta-learning on data following the \ptw{} prior.
        Middle: Per-time-step regret (see \cref{sec:methods}) measures the prediction error by quantifying the excess log-loss compared to a predictor that always knows the ground-truth bias.
        Bottom: Cumulative regret; the value at the final time-step is the basis for our main performance metric (see \cref{eq:cum_redundancy}).
    }
    \label{fig:single_trajectory}
\end{figure*}

In this work, we investigate the potential of MBML to learn parametric models that implicitly perform Bayesian inference with respect to more elaborate distributions than the ones investigated in \citet{mikulik2020meta}. We focus on \emph{piecewise stationary} Bernoulli distributions, which produce sequences that consist of Bernoulli \emph{segments} (see \cref{fig:single_trajectory}). The predictor only observes a stream of samples ($0$s and $1$s), with abrupt changes to local statistics at the unobserved switching-points between segments.
The focus on piecewise stationary sources is inspired by natural language, where documents often switch topic without explicit indication \citep{xie2022explanation}, and observation-action streams in environments with discrete latent variables, \eg multi-task RL without task-indicators.
In both domains, neural models that minimize sequential prediction error demonstrate hallmarks of sequential Bayesian prediction: strong context sensitivity or ``in-context learning'' \citep{reed2022generalist}, and rapid adaptation or ``few-shot learning'' \citep{brown2020language}.

To solve the sequential prediction problem, Bayes-optimal (BO) predictors simultaneously consider a number of hypotheses over switching-points and use prior knowledge over switching-points and segment-statistics.
Tractable exact BO predictors require non-trivial algorithmic derivations, and are only known for certain switching-point distributions.
The main question of this paper is whether neural predictors with memory, trained by minimizing sequential prediction error (log loss), can learn to mimic Bayes-optimal solutions and match their prediction performance.

Our contributions are:
\begin{itemize}
    \item Review of the theoretical connection between minimizing sequential prediction error, meta-learning, and its implied Bayesian objective (\cref{sec:mbml}).
    \item Theoretical argument for the necessity of memory to minimize the former (Bayesian) objective (\cref{sec:role_of_memory}).
    \item Empirical demonstration that meta-learned neural predictors can match prediction performance of two general non-parametric Bayesian predictors (\cref{sec:results}).
    \item Comparison of off-distribution generalization of learned solutions and Bayesian algorithms (\cref{sec:results}).
    \item Source code available at: \url{https://github.com/deepmind/nonstationary_mbml}.
\end{itemize}

    \section{Background}

We begin with some terminology for sequential, probabilistic data generating sources.
An alphabet is a finite, non-empty set of symbols, which we denote by $\cX$. 
A string $x_1x_2 \ldots x_n \in \cX^n$ of length $n$ is denoted by $x_{1:n}$.
The prefix $x_{1:j}$ of $x_{1:n}$, $j\leq n$, is denoted by $x_{\leq j}$ or $x_{< j+1}$.
The empty string is denoted by $\epsilon$.
Our notation also generalizes to out of bounds indices; that is, given a string $x_{1:n}$ and an integer $m > n$, we define $x_{1:m} := x_{1:n}$ and $x_{n:m}:=\epsilon$.
The concatenation of two strings $s$ and $r$ is denoted by $sr$.

\paragraph{Probabilistic Data Generating Sources}
A probabilistic data generating source $\rho$ is defined by a sequence of probability mass functions $\rho_n : \cX^n \to [0,1]$, for all $n\in\mathbb{N}$, satisfying the compatibility constraint that 
$\rho_n(x_{1:n}) = \sum_{y\in\cX} \rho_{n+1}(x_{1:n}y)$
for all $x_{1:n} \in \cX^n$, with base case $\rho_0(\epsilon) = 1$.
From here onward, whenever the meaning is clear from the argument to $\rho$, the subscripts on $\rho$ will be dropped.
Under this definition, the conditional probability of a symbol $x_n$ given previous data $x_{<n}$ is defined as $\rho(x_n | x_{<n}) := \rho(x_{1:n}) / \rho(x_{<n})$ provided $\rho(x_{<n}) > 0$, with the familiar chain rules $\rho(x_{1:n}) = \prod_{i=1}^n \rho(x_i | x_{<i})$ and $\rho(x_{i:j} \cdbar x_{<i}) = \prod_{k=i}^j \rho(x_k | x_{<k})$ now following.

\paragraph{Temporal Partitions}

A sub-sequence is described via a segment, which is a tuple of time-indices $(a,b) \in \mathbb{N}\times\mathbb{N}$ with $a \leq b$.
A segment $(a,b)$ is said to overlap with another segment $(c,d)$ if there exists an $i \in \mathbb{N}$ such that $a \leq i \leq b$ and $c \leq i \leq d$.
Let $S = \{ 1,2,\dots\,n \}$ denote a set of time-indices for some $n\in\mathbb{N}$.
A temporal partition $\cP$ of $S$ is a set of non-overlapping segments such that each $i\in S$ is covered by exactly one segment $(a,b) \in \cP$ with $a \leq i \leq b$.
We also use the overloaded notation $\cP(a,b) := \{ (c,d) \in \cP \;:\; a \leq c \leq d \leq b \}$.
Finally, $\cT_n$ will be used to denote the set of all possible temporal partitions of $\{ 1, 2, \dots, n \}$.

\paragraph{Piecewise Stationary Sources}

We now define a piecewise stationary data generating source $\mu$ in terms of a partition $\cP = \left\{ (a_1,b_1), (a_2,b_2), \dots \right \}$ and a set of probabilistic data generating sources $\{ \mu^1, \mu^2, \dots \},$ such that for all $n \in \mathbb{N}$, for all $x_{1:n} \in \cX^n$,
\begin{equation}
\label{eq:piecewise_source}
    \mu(x_{1:n}) := \prod_{(a,b)\in\cP_n} \mu^{f(a)}(x_{a:b}),
\end{equation}
where $\cP_n := \left \{ (a_i, b_i) \in \cP \,:\, a_i \leq n \right \}$ and $f(i)$ returns the index of the time segment containing $i$; that is, it gives a value $k \in \mathbb{N}$ such that both $(a_k, b_k) \in \cP$ and $a_k \leq i \leq b_k$.
In other words: a piecewise stationary data generating source consists of a number of non-overlapping segments (covering the entire range without gaps), with one stationary data generating distribution per segment.
An example-draw from such a source is shown in \cref{fig:single_trajectory}, where the distribution per segment is a Bernoulli process.

    \section{Memory-Based Meta-Learning}
\label{sec:mbml}

Given a parametric, memory-dependent probabilistic model $\rho_\theta(x_{1:n})$, a standard MBML setup works by repeating the following steps:
\begin{enumerate}
    \item Sample a task $\tau$ from a task distribution $\psi$;
    \item Generate data $x_{1:n} \sim \tau$; 
    \item Perform one or more steps of optimization of the model parameters $\theta$ using the loss $-\log \rho_\theta(x_{1:n}) = - \sum_{i=1}^n \rho_\theta(x_i \cdbar x_{<i})$.
\end{enumerate}

In our piecewise stationary Bernoulli setup, a task corresponds to prediction on a particular binary sequence (meaning $\tau$ is an instance of switching-points and Bernoulli biases for each segment), and the distribution over tasks is exactly the piecewise stationary distribution.
In the case where the task distribution is defined over a finite number of tasks, the marginal probability of the MBML data generating source is simply:
\begin{equation}
\label{eq:bayes_mbml}
    \xi(x_{1:n}) = \sum_{\tau} \psi(\tau) \, \tau(x_{1:n}).
\end{equation}
In other words: in meta-learning, the training data is \emph{implicitly} generated by a Bayesian mixture whose properties are determined from the particular details of the meta-training setup.
Note that this marginal form of a Bayesian mixture still captures the usual notion of  posterior updating implicitly; see \cref{sec:a_bayesmix} for more background.

\paragraph{Optimality of Bayesian Predictor for MBML}

Consider the expected excess log loss of using any sequential predictor $\rho$ on data $x_{1:n} \sim \xi$.
Notice that for all $n \in \mathbb{N}$, we have that
\begin{align}
\label{eq:kl_xi_rho}
    \mathbb{E}_{\xi}\left[-\log \rho(x_{1:n}) + \log \xi(x_{1:n})\right] = \nonumber \\ 
    \mathbb{E}_{\xi}\left[\log \frac{\xi(x_{1:n})}{\rho(x_{1:n})}\right] = D_{KL}(\xi \, || \, \rho) \geq 0,
\end{align}
with equality holding if and only if $\rho = \xi$ by the Gibbs inequality. 

In the context of our the generic MBML setup, \cref{eq:kl_xi_rho} implies that the Bayesian mixture $\rho=\xi$ (as given by \cref{eq:bayes_mbml}) is the unique optimal predictor in expectation. The set of all hypotheses/tasks in the mixture is called the model class $\mathcal{M}$.
Neural networks trained to minimize log loss should thus converge towards the Bayes-optimal solution (see \citet{ortega2019meta} for a detailed theoretical analysis).
Two conditions need to be fulfilled for trained meta-learners to behave Bayes-optimally:
\begin{enumerate}
    \item Realizability: the amortized Bayes-optimal solution needs to be representable by the model with the right set of parameters.
    \item Convergence: training needs to converge to this set of parameters.
\end{enumerate}
The hope is that by using sufficiently powerful function approximation techniques such as modern neural network architectures in an MBML setup, we can circumvent the need for explicit Bayesian inference and instead get the computational advantages associated with the Bayes-optimal predictor from a learned model with fixed weights.
But what properties of a model are needed for it to be sufficiently powerful?
The next section formally shows the necessity of using models with memory to achieve the Bayesian ideal. After establishing theoretically that Bayes-optimal predictors require memory, it is far from clear that memory-based neural network architectures achieve realizability (\ie have a set of parameters that represents the Bayes-optimal predictor) and convergence (via mini-batch based SGD).
We investigate these questions empirically in \cref{sec:results}.

    \section{The Essential Role of Memory}
\label{sec:role_of_memory}

It is important to emphasize that a fixed-parametric \mbox{\emph{memoryless}} model cannot, in general, learn the Bayesian mixture predictor~$\xi$ (with model class $\mathcal{M}$). The intuition is that the Bayesian mixture requires computation of posterior mixture weights, which, in general, depend on the history observations (the sufficient statistics) and thus necessitate some form of memory.
We now state this formally.

\begin{definition}
    A model $\nu$ is defined to be memoryless if $\nu$ can be written in the form $\nu_\Theta(x_{1:n}) := \prod_{i=1}^n \nu_{\theta_i}(x_i)$, where $\Theta = ( \theta_i )_{i=1}^n$ for all $x_{1:n}$. 
\end{definition}

In other words, $\nu_\theta$ is a product measure.
Next we present a negative result which explicitly quantifies the limitations of memoryless models to approximate general Bayesian inference.

\begin{theorem}
    Assume there exist $\mu_1,\mu_2\in\cM$ such that $\exists a_{1:\infty}:|\mathbb{E}_{\mu_1}[\mu_1(a_t|x_{<t})]-\mathbb{E}_{\mu_2}[\mu_2(a_t|x_{<t})]|\not\rightarrow 0$.
    Then there does not exist a $\Theta =(\theta_t)_{t=1}^\infty$ for a memoryless model $\nu_\Theta$ such that for all $\mu\in\cM$ we have $\mathbb{E}_{\mu}|\nu_\Theta(a_t|x_{<t})- \xi(a_t|x_{<t})|\to 0$ as $t\to\infty$.
    \label{thm:limitations}
\end{theorem}
For instance, for $\mu_i=\text{Bernoulli}(\vartheta_i)$, which are in most classes $\cM$, 
we have $|\mathbb{E}_{\mu_1}[\mu_1(a_t|x_{<t})]-\mathbb{E}_{\mu_2}[\mu_2(a_t|x_{<t})]|=|\vartheta_1-\vartheta_2|\neq 0$
for any choice of $\vartheta_1\neq\vartheta_2$.

The main intuition is that a discrete Bayesian mixture cannot always be represented as a product measure, as $\xi(x_n \cdbar x_{<n}) = \sum_{\rho  \in \cM} w^{\rho}_{n-1} \rho(x_n \cdbar x_{<n})$, where the posterior weight $w^{\rho}_{n-1} := w^\rho_0 \, \rho(x_{<n}) / \xi(x_{<n})$ for $n > 1$; in other words, $w^\rho_{n-1}$ can depend upon the whole history. 
A complete proof is given in \cref{subsec:proof-4.2}.

Importantly, this argument is independent of the representation capacity of $\nu_{\theta}$, and for example still holds even if $\nu_{\theta}$ is a universal function approximator, or if $\nu_{\theta}$ can represent each possible $\rho \in \cM$ given data \emph{only} from $\rho$. 
The same argument extends to any $k$-Markov stationary model for finite $k$, though one would expect much better approximations to be possible in practice with larger $k$.

    \section{Priors and Exact Inference Baselines}
\label{sec:priors_and_baselines}

This section describes our baseline Bayesian algorithms for exact Bayesian inference on piecewise stationary Bernoulli data.
The algorithms make different assumptions regarding the statistical structure of switching-points.
If the data generating source satisfies these assumptions, then the baselines are theoretically known to perform optimally in terms of expected cumulative regret.
This allows us to assess the quality of the meta-learned solutions against known optimal predictors.
Note that while exact Bayesian inference is often computationally intractable, the cases we consider here are noteworthy in the sense that they can be computed efficiently, and in some cases with quite elaborate algorithms involving combinations of dynamic programming (see \citet{koolen2008combining} for a comprehensive overview) and the generalized distributive law \citep{aji2000generalized}.

In order to ensure that the data generating source matches the statistical prior assumptions made by the different baselines, we use their underlying priors as data generating distributions in our experiments (see \cref{sec:prior_sampling} for details on the algorithms that sample from the priors). 

\paragraph{KT Estimator}

The KT estimator is a simple Beta-Binomial model which efficiently implements a Bayesian predictor for $\text{Bernoulli}(\theta)$ sources with unknown $\theta$ by maintaining sufficient statistics in the form of counts.
By using a $\text{Beta}(\tfrac{1}{2},\tfrac{1}{2})$ prior over $\theta$, we obtain the KT-estimator \citep{krichevsky1981performance}, which has optimal worst case regret guarantees with respect to data generated from an unknown Bernoulli source.
Conveniently, the predictive probability has a closed form
\begin{equation*}
    \text{\sc kt}(x_{n+1} = 1 \cdbar x_{1:n}) = \frac{c(x_{1:n}) + \tfrac{1}{2}}{n+1},
\end{equation*}
where $c(x_{1:n})$ returns the number of ones in $x_{1:n}$, and $\text{\sc kt}(x_{n+1} = 0 \cdbar x_{1:n}) = 1 - \text{\sc kt}(x_{n+1} = 1 \cdbar x_{1:n})$.
This can be implemented efficiently online by maintaining two counters,
and the associated marginal probability can be obtained via the chain rule $\text{\sc kt}(x_{1:n}) = \prod_{i=1}^n \text{\sc kt}(x_{i}  \cdbar x_{<i}) $.
The KT estimator cannot handle (piecewise) non-stationary distributions; to allow for this we next make a simple extension, and later more complex extensions.

\paragraph{KT Oracle}

Our first baseline extends the KT estimator to deal with piecewise stationarity: KT Oracle is provided with knowledge of when switching-points occur.
This allows using a KT estimator and simply resetting its counters at each switching-point.
The KT Oracle serves as a lower bound to show achievable regret in case switching-points could be instantaneously predicted with perfect accuracy.
The prior underlying the KT Oracle is never used to generate data in our experiments, since the KT Oracle does not specify a distribution over switching-points.

\paragraph{\ptw: Partition Tree Weighting}

Our second baseline is Partition Tree Weighting \citep{veness2013partition}.
In contrast to the KT Oracle, \ptw{} does not need to observe switching-points.
Instead, it performs Bayesian model averaging over a carefully chosen subset $\cC_d \subset \cT_n$ of temporal partitions by computing 
\begin{equation*}
    {\normalfont\textsc{ptw}}_d(x_{1:n}) = \sum_{\cP \in \cC_d} 2^{-\Gamma_d(\cP)} \prod_{(a,b) \in \cP} \rho(x_{a:b}),
\end{equation*}
where $\rho$ is a base-predictor for a single segment (in our case the KT-estimator), and $d$ is the depth of the partition tree which needs to be at least $\log n$.
In other words, the technique gives a way to extend a given base predictor $\rho$ to a piecewise setting, with known worst case regret guarantees that follow from the use of model averaging over a tree structured prior.
Although the number of partitions $|\cC_d|$ grows $O(2^{2^d})=O(2^n)$, this technique adds only a $O(\log n)$ time/space overhead compared with computing $\rho(x_{1:n})$, and can be computed online in a recursive/incremental fashion.
In this work we restrict our attention to the case where the base model is the KT-estimator, $\rho = \kt$, to obtain a low-complexity universal algorithm for piecewise Bernoulli sources.
Informally, \ptw{} assumes that a trajectory has a switching-point at half its length with probability $1/2$, and both resulting sub-trajectories also have a switching-point at their respective halves with probability $1/2$, and so on (recursively) for all subsequent sub-trajectories.
This assumption allows for efficient implementation and leads to a characteristic inductive bias.
In our experiments we investigate whether neural models can meta-learn this structured inductive bias and match prediction performance of \ptw{} on data that follows these assumptions.

\paragraph{\lin: Exact Model Averaging Over All Temp. Partitions}

Our final baseline, \lin, is the linear complexity method introduced by \citet{willems1996coding}.
It performs Bayesian model averaging over all temporal partitions (whereas \ptw{} only considers a subset), and all possible Bernoulli models within each segment, and has the marginal form
\begin{equation*}
    {\normalfont\textsc{lin}}(x_{1:n}) = \sum_{\cP \in \cT_n} w(\cP) \prod_{(a,b) \in \cP} \kt(x_{a:b}),
\end{equation*}
where $w(\cP)$ is a prior over the linear-transition diagram representation of $\cP$, the details of which are not important for this work, but they introduce a different assumption over the distribution and location of switching-points compared to \ptw.
To process a sequence of $n$ symbols, this algorithm runs in time $O(n^2)$ and has space complexity of $O(n)$.
In our experimental section we also investigate whether neural models can meta-learn to match the inductive bias of \lin.

    \section{Methodology}
\label{sec:methods}

The general approach for our experiments is to train various memory-based neural models according to the MBML training setup described in \cref{sec:mbml}.
We explore multiple neural architectures to get a better sense as to how architectural features influence the quality of the meta-learned Bayesian approximation.
After training, we evaluate models either on data drawn from the same meta-distribution as during training (on-distribution experiments) or from a different distribution (off-distribution experiments).
We quantify prediction performance by the expected cumulative regret (called redundancy in information theory) with respect to the ground-truth piecewise data generating source $\mu$, quantifying the expected excess log loss of the neural predictor. 
More formally, we define the expected instantaneous regret of model $\pi$ at time $t$ with respect to the piecewise source $\mu$ as
\begin{equation*}
    R_{\pi\mu}(t) := \mathbb E_{x_t \sim \mu^{f(t)}} \left[ \log \mu^{f(t)}(x_t) - \log \pi (x_t) \right],
\end{equation*}
compare \cref{eq:kl_xi_rho}, and the cumulative expected regret as 
\begin{equation}
\label{eq:cum_redundancy}
    R_{\pi\mu}^T := \sum_{t=1}^T  R_{\pi\mu}(t).
\end{equation}
An illustration of both metrics is shown in \cref{fig:single_trajectory}.
Note that a cumulative expected regret of zero corresponds to the performance of an oracle which knows both the location of the switching-points, as well as the parameter of each Bernoulli process governing a segment.

We now introduce the different types of data generating sources used in our experiments, before describing the different types of memory-based neural models that we evaluated.

\begin{figure}[htb]
    \centering
    \includesvg[scale=0.6]{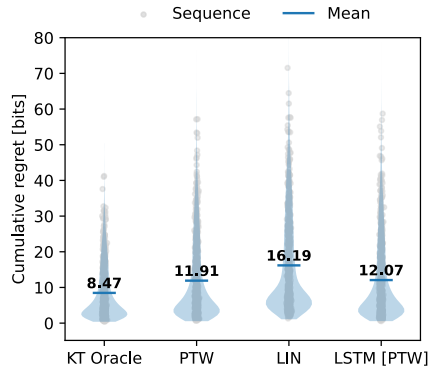}
    \caption{
        Mean cumulative regret across $10$k~sequences of length~$256$ drawn from \ptw{} prior (same setting as \cref{fig:single_trajectory}).
        The LSTM trained on data from the \ptw{} prior matches prediction performance of the optimal \ptw{} predictor.
        We also compare against \lin, a strong but suboptimal predictor for this distribution.
    }
    \label{fig:single_seed}
\end{figure}

\begin{figure*}[t!]
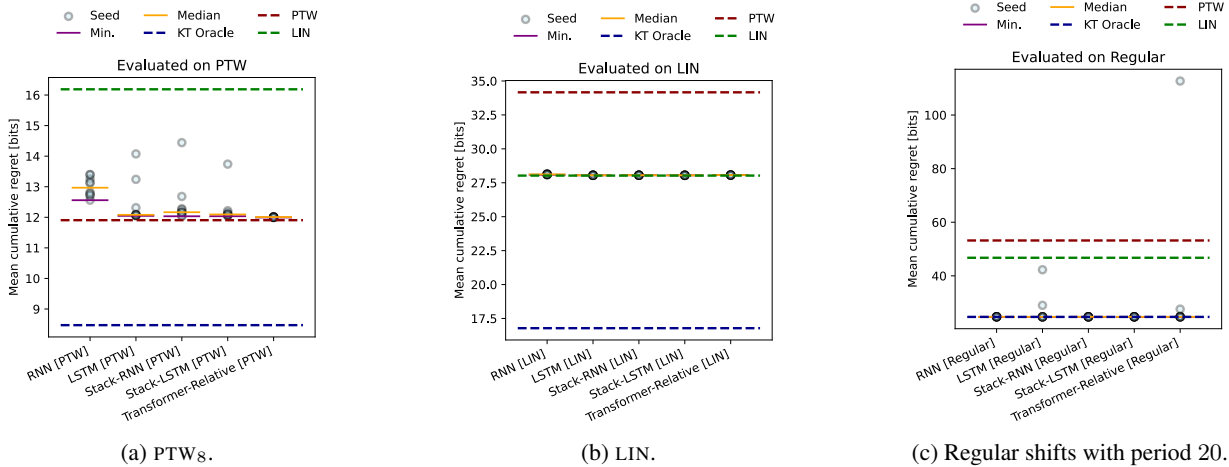

    \centering
    \begin{minipage}[t]{0.3\textwidth}
        \centering
         \includesvg[scale=0.45]{figures/main_result_ptw.svg}
         \linebreak \small{(a) \ptw$_8$.}
    \end{minipage}\hfill
    \begin{minipage}[t]{0.3\textwidth}
        \centering
        \includesvg[scale=0.45]{figures/main_result_lin.svg}
        \linebreak \small{(b) \lin.}
    \end{minipage}\hfill
    \begin{minipage}[t]{0.3\textwidth}
        \centering
        \includesvg[scale=0.45]{figures/main_result_regular.svg}
        \linebreak \small{(c) Regular shifts with period $20$.}
    \end{minipage}
    \caption{
        On-distribution performance (models trained and evaluated on same distribution, denoted below panels).
        Evaluation on $10$k sequences of length~$256$.
        Columns in each plot show individual trained models (circles), and minimum and median results across random initializations.
        Square-brackets denote the training distribution for models.
        Dashed lines show the three exact Bayesian inference algorithms as dashed lines---of course \ptw{} and \lin{} are only optimal for their respective data regimes, but serve as a strong baseline predictor in the other regimes.
    }
    \label{fig:main_result_on_distribution}
\end{figure*}

\paragraph{Data-Generation}

We consider data sources that are piecewise stationary in the form given by Equation~\eqref{eq:piecewise_source}.
Within a stationary segment $i$,  $\mu^i$ is  a Bernoulli distribution with bias sampled from a Beta prior $\mu^i \sim \text{Beta}(\alpha, \beta)$; see \cref{fig:single_trajectory} for a concrete example. 
In our experiments, we always use $\alpha = \beta = 0.5$, which is consistent with the prior used by the KT-estimator. 

Across our experiments, we consider four different distributions over switching-points, two of which coincide with the statistical assumptions of our exact inference baselines (\ptw{} and \lin): 
\begin{itemize}
    \item \textbf{Regular Periodic:}
    All segments have fixed length $l$, meaning that switching-points occur deterministically at the same locations across all sampled trajectories. Neural predictors can, during meta-learning, pick up on $l$ and thus learn to predict switching-points with perfect accuracy.
    \item \textbf{Random Uniform:}
    Segment-lengths are repeatedly drawn from a $\text{Uniform}(1, n)$ distribution until the combined summed segment length matches or exceeds the desired sequence length $n$.
    \item \textbf{\ptw{} prior:}
    Switching-points are sampled from the \ptw{} prior.
    More specifically, a temporal partition can be sampled from the \ptw$_d$ prior using \cref{alg:ptw_prior_sample} with an expected running time of $O(d)$, where $d$ is the depth of the partition-tree; see \cref{sec:prior_sampling} for more detail.
    Unless otherwise indicated, \ptw{} in our experiments refers to using the minimally necessary depth for the given sequence length, \eg \ptw$_8$ for length~$256$ and \ptw$_9$ for length~$512$.
    \item \textbf{\lin{} prior:}
    Switching-points are sampled from the \lin{} prior.
    \Cref{alg:lin_prior_sample} in \cref{sec:prior_sampling} provides a method for sampling temporal partitions from the \lin{} prior, whose worst-case time and space complexity grows linearly with the sequence length $n$. 
\end{itemize}
Example draws and visualizations of the switching-point statistics of all prior distributions are shown in \cref{sec:example-sequences}.

\paragraph{Neural Predictors}

Our neural models sequentially observe binary samples from the data generating source and output probabilities over the next observation.
$ \pi_\theta( \cdot | x_{<t})$  given their parameters $\theta$ and the data seen so far up to time $t$. 
We use the logarithmic loss for training;
for a sequence up to time $T$, we have $\ell_\theta (x_{1:T}):= -\frac{1}{T} \sum_{t=1}^T \log \pi_\theta (x_t \vert x_{<t})$.
During training, parameters are updated via mini-batch stochastic gradient descent using ADAM.

We evaluate the following network architectures:
\begin{itemize}
    \item \textbf{RNN:} One layer of vanilla RNN neurons, followed by a two-layer fully connected read-out.
    \item \textbf{LSTM:} One layer of LSTM~\citep{hochreiter1997long} memory cells, followed by a two-layer fully connected read-out.
    \item \textbf{Stack-RNN/LSTM:} We also augment the LSTM and RNN predictors with a stack, similar to the Stack-RNN of~\citet{joulin2015inferring}.
    The stack has three operations, \textsc{push}, \textsc{pop}, and \textsc{no-op}, which are implemented in a ``soft'' fashion for differentiability, \ie stack updates are computed via a linear combination of each stack-action probability.
    At each time-step the RNN/LSTM reads the top of the stack as an additional input.
    A push writes a lower-dimensional projection of the RNN/LSTM cell states to the top of the stack.
    We treat the dimensionality of the projection and the maximum depth of the stack as hyperparameters.
    \item \textbf{Transformer:} We use a Transformer encoder with incremental causal masking to implement sequential online prediction.
    The context of the transformer thus acts as a (verbose) memory, storing all observations seen so far.
    In our ablations we also simulate having a smaller context length (via masking), but the best results are achieved with the full context.
    We evaluate three different positional encodings (see \cref{sec:ablation-study}): standard sin/cos~\cite{vaswani2017attention}, ALiBi~\cite{press2022train}, and the relative positional encodings from TransformerXL~\cite{dai2019transformer}.
    For our experiments in \cref{sec:results}, we use the relative encoding, as it performed best in the ablations.
\end{itemize}

For all our network architectures, we conducted an initial ablation study to determine architecture hyperparameters (see \cref{sec:ablation-study}).
The experimental results shown in \cref{sec:results} use the hyperparameter-set that led to the lowest expected cumulative redundancy in the ablations (we provide the exact values in \cref{sec:ablation-study}). 

We provide an open-source implementation of our models, tasks, and training and evaluation suite at \url{https://github.com/deepmind/nonstationary_mbml}.

    \section{Results}
\label{sec:results}

To clarify how our main results are computed, an example sequence from a \ptw{} source, and corresponding model predictions, as well as our performance metric, are shown in \cref{fig:single_trajectory}; example draws from the other sources are in \cref{sec:example-sequences}.
To compare models' performance we empirically compute the mean cumulative regret across $10$k sequences, see \cref{fig:single_seed}. 
Finally, we perform the same evaluation over $10$ different random initializations for each model.

\paragraph{On-Distribution Evaluation}

We first evaluate the performance of neural models when trained and evaluated on the same data generating distribution---results shown in \cref{fig:main_result_on_distribution}.
Generally, we find that neural models match prediction performance of the Bayes-optimal predictors very well on their respective data regimes.
Picking the best random initialization (Min in the figure), all neural predictors achieve near-optimal performance, except the RNN which has a slightly larger error on the \ptw{} data.
Median results (across random initializations) reveal some differences in training stability.
It is quite remarkable that all neural models across all random seeds, when trained on \lin{} data, manage to match \lin{} performance almost exactly.
Somewhat less surprising, for regular periodic shifts all neural models quite reliably learn to predict switching-points with perfect accuracy, allowing them to reach KT Oracle performance levels.
\cref{fig:main_result_random} in the Appendix shows on-distribution evaluation results for the Random Uniform distribution.

\paragraph{Off-Distribution Evaluation}

\begin{figure}
    \centering
    \includesvg[scale=0.5]{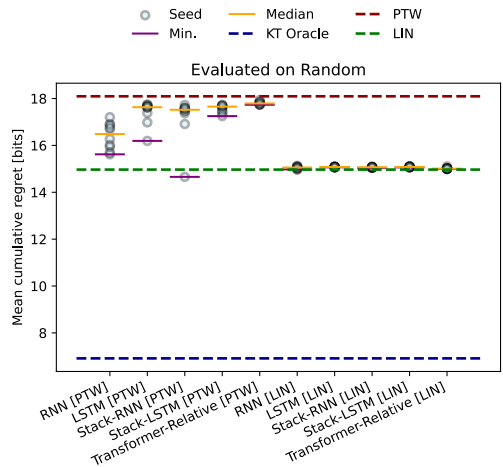}
    \caption{
        Off-distribution evaluation ($10$k sequences, length~$256$).
        Models' training distribution indicated in the square brackets.
        All models are evaluated with a random-uniform distribution over segment lengths ($\text{Uniform}(1, 256)$).
        Red dashed line shows \ptw$_8$.
    }
    \label{fig:generalisation_random}
\end{figure}

The experiments in this section serve to illustrate that models pick up precise inductive biases during meta-learning.
Biases, that match the statistical structure of the data distribution during training.
If the data distribution at test time violates this statistical structure, optimal prediction performance can no longer be guaranteed.
\cref{fig:generalisation_random} shows how models trained on data from the \ptw{} and \lin{} prior perform when evaluated with data drawn from a random uniform changepoint distribution. 
Overall, neural networks trained on \ptw{} are slightly more robust against this change compared to \ptw---the better neural models fit \ptw{} in \cref{fig:main_result_on_distribution} (a), the less robust they seem to be against this distributional shift.
Off-distribution generalization for the models trained on \lin{} is very uniform across models and closely aligned with the exact inference implementation in terms of prediction performance.
We show more off-distribution evaluations in \cref{sec:off-distribution-generalization}. 

\paragraph{Sequence-Length Generalization}

\begin{figure}
    \centering
    \includesvg[scale=0.5]{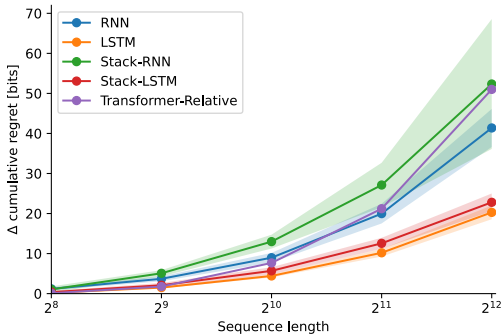}
    \caption{
        Evaluation of models on longer sequences.
        Models are trained on length~$256$ with switching-points drawn from \ptw$_8$ (same as \cref{fig:main_result_on_distribution} (a)) and evaluated on sequences up to length~$4096$ (depth of \ptw{} is $\log_2$(sequence~length)).
        The plot shows the difference between the models' cumulative regret and \ptw{} over $1$k sequences.
        Lines show the mean and shaded areas the standard deviation over 10 random seeds.
        The LSTM and Stack-LSTM generalize best, but for all models performance degrades as the sequence length increases beyond the training length, which is a signature of learned amortized inference.
    }
    \label{fig:generalisation-longer-lengths}
\end{figure}

\cref{fig:generalisation-longer-lengths} shows length-generalization behavior of the neural models.
All models shown are trained on sequences of length~$256$ but evaluated on much longer sequences.
As expected the models' performance degrades with longer sequences, but remains reasonably good, indicating that, \eg internal dynamics of the recurrent networks do not break down catastrophically.
See \cref{fig:single_traj_generalisation_ptw_lstm_512} for an example trajectory for the LSTM evaluated on a sequence of length~$512$, showing that predictions overall remain quite close to the optimum. 

\begin{figure}
    \centering
    \includesvg[scale=0.5]{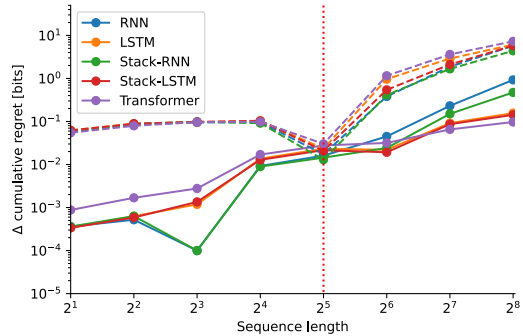}
    \caption{
        Evaluation of models on sequences of different lengths. The plot shows the difference between the models' expected cumulative regret and \ptw{} over $1$k sequences (depth of \ptw{} is $\log_2$(sequence~length)). Results are averaged over $10$ random seeds.
        Solid lines correspond to models evaluated on the length they were trained on.
        Dashed lines correspond to models trained on length~$32$ (dotted vertical line) and evaluated on other lengths.
        As expected, models trained on $32$ generalize worse to other lengths ('U' shape curve), which is explained by the implicit distributional shift induced by the \ptw{} prior with different depth.
    }
    \label{fig:generalisation-other-lengths}
\end{figure}

Note that the most likely switching-points under the \ptw{} prior depend on the sequence length, and thus our sequence-length generalization experiment also induces a slight distributional shift (models trained on length $256$ have a different prior expectation over switching point locations than the \ptw{} prior assigns for shorter or longer sequence lengths).
To quantify this effect \cref{fig:generalisation-other-lengths} shows results of a sequence-length ablation that compares two types of models: one, models trained on length~$32$ and evaluated on shorter and longer lengths (suffering from the implicit distributional shift that arises from \ptw{} priors of different depth), and two, models evaluated on the length that they were trained on (for a range of different lengths).

    \section{Related Work and Discussion}
\label{sec:related-work}

Meta-learning is a technique for producing data-efficient learners at test time through the acquisition of inductive biases from training data \citep{bengio1991learning, schmidhuber1996simple, thrun1998learning}.
Recently, \citet{ortega2019meta} showed theoretically how (memory-based) meta-learning leads to predictors that perform amortized Bayesian inference, \ie meta-learners are trained to minimize prediction error (log loss) over a task distribution which requires (implicit) inference of the task at hand. 
Minimal error is achieved by taking into account a priori regularities in the data in a Bayesian fashion and, in decision-making tasks, implies automatically trading-off exploration and exploitation \citep{zintgraf2020varibad}.
Memory-based meta-learners pick up on a priori statistical regularities simply by training over the distribution of tasks without directly observing task indicators.
This leads to parametric functions that implement amortized Bayesian inference \citep{gershman2014amortized, ritchie2016deep}, where a parametric model~$\pi_\theta$ behaves as if performing Bayesian inference ``under the hood'': $\pi_\theta(x_{<t}) \approx p(x_t \vert x_{<t}) = \sum_\tau p(x_t \vert \tau, x_{<t}) p(\tau \vert x_{<t})$. 
The r.h.s. requires posterior inference over the task-parameters~$p(\tau \vert x_{<t}) \propto p(x_{<t} \vert \tau) p(\tau)$, which is often analytically intractable. 
The result is a model with fixed parameters that implements an adaptive algorithm via its activations, and at its core is the collection of sufficient statistics for rapid online task inference. 
The argument can be extended to Bayes-optimal decision-making \citep{ortega2019meta, mikulik2020meta}; recently, \citet{bauer2023human} reported a large-scale demonstration of the principle, where models are trained over $25$ billion distinct tasks in simulated 3D environments. 
Trained models are able to adapt to novel tasks on human time-scale (\ie with tens or a few hundreds of seconds of interaction) purely via in-context learning (conditioning). 
\citet{kirsch2022general} also conducted an exploration of memory-based meta-learning over a vast set of tasks to produce in-context and few-shot learning abilities, with up to $2^{24}$ tasks created by randomly projecting inputs and randomly permuting labels on MNIST.
They find that having both, a large enough model and a rich enough training distribution is required for an in-context learning algorithm that generalizes.

While Bayes-optimality in sequential prediction and decision-making is theoretically well understood, \cf \citet{hutter2005universal}, an important question is whether neural networks, when meta-trained appropriately, can approach the Bayesian solution at all (realizability and convergence, see \cref{sec:mbml}), or whether they operate primarily in a suboptimal regime that is not well described by Bayesian theory.
\citet{mikulik2020meta} conducted a first targeted empirical comparison of meta-learned neural predictors with Bayes-optimal algorithms, focusing on simple prediction- and decision-making tasks where episodes had a fixed number of steps, and changepoints were observed (internal memory states were explicitly reset at episode boundaries).
This setting is similar to our regular periodic switching-point distribution, but, crucially, switching-points are always unobserved in our experiments.
That is, the emphasis of our study is on non-stationary data sources with abrupt changes in local statistics.
While piecewise stationary sources are conceptually simple, the switching-points make accurate prediction challenging, particularly under a cumulative error metric.
Furthermore, piecewise Bernoulli data makes switching-point detection difficult, which is, counter-intuitively, often easier on more complex distributions when different segments exhibit strongly characteristic statistics.
In \citet{reed2022generalist}, observations are, for instance, frames from Atari games, where a single frame often suffices to determine the task accurately. 

We also aim at furthering the understanding of inductive biases and reasoning principles acquired by sequential predictors such as large language models.
Recently observed in-context learning abilities in large language models \citep{brown2020language} have rekindled interest in black-box parametric models capable of learning-to-learn purely in-context, that is, via activations, with frozen parameters \citep{hochreiter2001learning, duan2016rl, santoro2016meta, wang2017learning}.
While the capabilities to learn in-context have been heavily explored empirically, the connections to Bayesian theory are still somewhat sparse \citep{ortega2019meta, mikulik2020meta, muller2021transformers, xie2022explanation}.
From an AI safety viewpoint it is desirable to understand the mechanisms that enable few-shot and in-context learning; which are plausibly the same mechanisms that create susceptibility to prompt injections and context poisoning attacks.
These characteristics are expected from a model that performs implicit Bayesian inference over piecewise stationary data.
For instance, \citet{xie2022explanation} argued that in-context learning in large language models can be explained by (implicit) Bayesian inference over a latent variable, but does not draw a connection to the theory of meta-learning (which explains why amortized Bayesian inference arises from minimizing log loss) and does not compare against a known Bayes-optimal algorithm to establish optimality of the neural predictor.
Our meta-learning interpretation is in line with the arguments in \citet{xie2022explanation} but is more general. Our interpretation also does not rely on special delimiter characters that signal a topic switch and needing to have a posterior over the latent variable that is highly concentrated on a single value.
We believe it could be interesting in the future to contrast the meta-learning interpretation with the model by \citet{xie2022explanation} and extend our experimental suite to incorporate their hidden Markov model as a more complex piecewise stationary source.

\paragraph{Limitations}
Our results show the potential of memory-based meta-learning to accurately approximate Bayes-optimal solutions.
However, our findings are currently limited to Bernoulli statistics per segment, and four types of switching-point distributions.
For known Bayes-optimal algorithms the complexity of dealing with different switching-point distributions seems to dominate over increasing the complexity of the base distributions per segment. This makes us optimistic that our findings would generalize to more complex per-segment distributions when training neural predictors---but at the current stage this remains speculative.
The main challenge with more complex data generating sources, such as real-world datasets, is the lack of a (computationally or analytically) tractable Bayes-optimal solution against which we could compare. The main point of this paper is to demonstrate that neural networks can learn to predict Bayes-optimally and not simply that they can learn to predict well (which has already been demonstrated extensively in the literature).
Another limitation of our study is that many known Bayes-optimal algorithms come with performance guarantees and robustness bounds, and while our generalization experiments attempt to shed some light on robustness and out-of-distribution behavior of meta-learned neural models, no formal guarantees can be provided.

    \section{Conclusion}
\label{sec:conclusion}

In this paper we investigated whether neural networks, trained to minimize sequential prediction error (log loss) over statistically structured but highly non-stationary data sources, can learn to match the prediction performance of Bayes-optimal algorithms.
We found this to be the case, despite non-trivial algorithmic requirements for optimal prediction in these settings.
Our results empirically confirm the theoretical Bayesian interpretation of memory-based meta-learning \citep{ortega2019meta}, which states that log-loss minimization on a meta-distribution over data sources with a memory-based parametric model leads to approximately Bayes-optimal solutions.
By focusing on piecewise stationary data sources, we study a highly relevant regime that holds the promise to shed light onto recently observed capabilities of large sequential prediction models.
We believe that few-shot and in-context learning abilities of these models, as well as their susceptibility to context-corruption and prompt-injection attacks at test time, can be better understood from the viewpoint of inferring changes in local statistics under a non-stationary distribution. 
A more concrete, and near-term take-away from our study is to highlight the potential of using memory-based meta-learning to \emph{learn} (near-) Bayes-optimal predictors in settings where closed-form solutions are not obtainable or algorithmically intractable. The ingredients to succeed with this are highly expressive parametric models (for realizability of the Bayes-optimal predictor) and strong optimizers (to ensure convergence)---our current study shows that modern neural networks in a standard meta-learning setup with mini-batch based SGD can fit this bill.

    \section*{Acknowledgements}

We thank Jane Wang, Christopher Mattern, and Shane Legg for their helpful feedback and insightful conversations.

    \bibliography{references}

\begin{thebibliography}{29}
\providecommand{\natexlab}[1]{#1}
\providecommand{\url}[1]{\texttt{#1}}
\expandafter\ifx\csname urlstyle\endcsname\relax
  \providecommand{\doi}[1]{doi: #1}\else
  \providecommand{\doi}{doi: \begingroup \urlstyle{rm}\Url}\fi

\bibitem[{Adaptive Agent Team} et~al.(2023){Adaptive Agent Team}, Bauer,
  Baumli, Baveja, Behbahani, Bhoopchand, Bradley{-}Schmieg, Chang, Clay,
  Collister, Dasagi, Gonzalez, Gregor, Hughes, Kashem, Loks{-}Thompson,
  Openshaw, Parker{-}Holder, Pathak, Nieves, Rakicevic, Rockt{\"{a}}schel,
  Schroecker, Sygnowski, Tuyls, York, Zacherl, and Zhang]{bauer2023human}
{Adaptive Agent Team}, Bauer, J., Baumli, K., Baveja, S., Behbahani, F. M.~P.,
  Bhoopchand, A., Bradley{-}Schmieg, N., Chang, M., Clay, N., Collister, A.,
  Dasagi, V., Gonzalez, L., Gregor, K., Hughes, E., Kashem, S.,
  Loks{-}Thompson, M., Openshaw, H., Parker{-}Holder, J., Pathak, S., Nieves,
  N.~P., Rakicevic, N., Rockt{\"{a}}schel, T., Schroecker, Y., Sygnowski, J.,
  Tuyls, K., York, S., Zacherl, A., and Zhang, L.
\newblock Human-timescale adaptation in an open-ended task space.
\newblock \emph{CoRR}, abs/2301.07608, 2023.

\bibitem[Aji \& McEliece(2000)Aji and McEliece]{aji2000generalized}
Aji, S.~M. and McEliece, R.~J.
\newblock The generalized distributive law.
\newblock \emph{{IEEE} Trans. Inf. Theory}, 46\penalty0 (2):\penalty0 325--343,
  2000.

\bibitem[Bengio et~al.(1991)Bengio, Bengio, and Cloutier]{bengio1991learning}
Bengio, Y., Bengio, S., and Cloutier, J.
\newblock Learning a synaptic learning rule.
\newblock In \emph{IJCNN-91-Seattle International Joint Conference on Neural
  Networks}, 1991.

\bibitem[Brown et~al.(2020)Brown, Mann, Ryder, Subbiah, Kaplan, Dhariwal,
  Neelakantan, Shyam, Sastry, Askell, Agarwal, Herbert{-}Voss, Krueger,
  Henighan, Child, Ramesh, Ziegler, Wu, Winter, Hesse, Chen, Sigler, Litwin,
  Gray, Chess, Clark, Berner, McCandlish, Radford, Sutskever, and
  Amodei]{brown2020language}
Brown, T.~B., Mann, B., Ryder, N., Subbiah, M., Kaplan, J., Dhariwal, P.,
  Neelakantan, A., Shyam, P., Sastry, G., Askell, A., Agarwal, S.,
  Herbert{-}Voss, A., Krueger, G., Henighan, T., Child, R., Ramesh, A.,
  Ziegler, D.~M., Wu, J., Winter, C., Hesse, C., Chen, M., Sigler, E., Litwin,
  M., Gray, S., Chess, B., Clark, J., Berner, C., McCandlish, S., Radford, A.,
  Sutskever, I., and Amodei, D.
\newblock Language models are few-shot learners.
\newblock In \emph{NeurIPS}, 2020.

\bibitem[Dai et~al.(2019)Dai, Yang, Yang, Carbonell, Le, and
  Salakhutdinov]{dai2019transformer}
Dai, Z., Yang, Z., Yang, Y., Carbonell, J.~G., Le, Q.~V., and Salakhutdinov, R.
\newblock Transformer-xl: Attentive language models beyond a fixed-length
  context.
\newblock In \emph{{ACL} {(1)}}, pp.\  2978--2988. Association for
  Computational Linguistics, 2019.

\bibitem[Duan et~al.(2016)Duan, Schulman, Chen, Bartlett, Sutskever, and
  Abbeel]{duan2016rl}
Duan, Y., Schulman, J., Chen, X., Bartlett, P.~L., Sutskever, I., and Abbeel,
  P.
\newblock Rl{\textdollar}{\^{}}2{\textdollar}: Fast reinforcement learning via
  slow reinforcement learning.
\newblock \emph{CoRR}, abs/1611.02779, 2016.

\bibitem[Gershman \& Goodman(2014)Gershman and Goodman]{gershman2014amortized}
Gershman, S. and Goodman, N.~D.
\newblock Amortized inference in probabilistic reasoning.
\newblock In \emph{CogSci}. cognitivesciencesociety.org, 2014.

\bibitem[Hochreiter \& Schmidhuber(1997)Hochreiter and
  Schmidhuber]{hochreiter1997long}
Hochreiter, S. and Schmidhuber, J.
\newblock Long short-term memory.
\newblock \emph{Neural Comput.}, 9\penalty0 (8):\penalty0 1735--1780, 1997.

\bibitem[Hochreiter et~al.(2001)Hochreiter, Younger, and
  Conwell]{hochreiter2001learning}
Hochreiter, S., Younger, A.~S., and Conwell, P.~R.
\newblock Learning to learn using gradient descent.
\newblock In \emph{{ICANN}}, volume 2130 of \emph{Lecture Notes in Computer
  Science}, pp.\  87--94. Springer, 2001.

\bibitem[Hutter(2005)]{hutter2005universal}
Hutter, M.
\newblock \emph{Universal Artificial Intelligence: Sequential Decisions Based
  on Algorithmic Probability}.
\newblock Springer, 2005.

\bibitem[Joulin \& Mikolov(2015)Joulin and Mikolov]{joulin2015inferring}
Joulin, A. and Mikolov, T.
\newblock Inferring algorithmic patterns with stack-augmented recurrent nets.
\newblock In \emph{{NIPS}}, pp.\  190--198, 2015.

\bibitem[Kirsch et~al.(2022)Kirsch, Harrison, Sohl{-}Dickstein, and
  Metz]{kirsch2022general}
Kirsch, L., Harrison, J., Sohl{-}Dickstein, J., and Metz, L.
\newblock General-purpose in-context learning by meta-learning transformers.
\newblock \emph{CoRR}, abs/2212.04458, 2022.

\bibitem[Koolen \& de~Rooij(2008)Koolen and de~Rooij]{koolen2008combining}
Koolen, W.~M. and de~Rooij, S.
\newblock Combining expert advice efficiently.
\newblock In \emph{{COLT}}, pp.\  275--286. Omnipress, 2008.

\bibitem[Krichevsky \& Trofimov(1981)Krichevsky and
  Trofimov]{krichevsky1981performance}
Krichevsky, R.~E. and Trofimov, V.~K.
\newblock The performance of universal encoding.
\newblock \emph{{IEEE} Trans. Inf. Theory}, 27\penalty0 (2):\penalty0 199--206,
  1981.

\bibitem[Mikulik et~al.(2020)Mikulik, Del{\'{e}}tang, McGrath, Genewein,
  Martic, Legg, and Ortega]{mikulik2020meta}
Mikulik, V., Del{\'{e}}tang, G., McGrath, T., Genewein, T., Martic, M., Legg,
  S., and Ortega, P.~A.
\newblock Meta-trained agents implement bayes-optimal agents.
\newblock In \emph{NeurIPS}, 2020.

\bibitem[M{\"{u}}ller et~al.(2022)M{\"{u}}ller, Hollmann, Pineda{-}Arango,
  Grabocka, and Hutter]{muller2021transformers}
M{\"{u}}ller, S., Hollmann, N., Pineda{-}Arango, S., Grabocka, J., and Hutter,
  F.
\newblock Transformers can do bayesian inference.
\newblock In \emph{{ICLR}}. OpenReview.net, 2022.

\bibitem[Ortega et~al.(2019)Ortega, Wang, Rowland, Genewein, Kurth{-}Nelson,
  Pascanu, Heess, Veness, Pritzel, Sprechmann, Jayakumar, McGrath, Miller,
  Azar, Osband, Rabinowitz, Gy{\"{o}}rgy, Chiappa, Osindero, Teh, van Hasselt,
  de~Freitas, Botvinick, and Legg]{ortega2019meta}
Ortega, P.~A., Wang, J.~X., Rowland, M., Genewein, T., Kurth{-}Nelson, Z.,
  Pascanu, R., Heess, N., Veness, J., Pritzel, A., Sprechmann, P., Jayakumar,
  S.~M., McGrath, T., Miller, K.~J., Azar, M.~G., Osband, I., Rabinowitz,
  N.~C., Gy{\"{o}}rgy, A., Chiappa, S., Osindero, S., Teh, Y.~W., van Hasselt,
  H., de~Freitas, N., Botvinick, M.~M., and Legg, S.
\newblock Meta-learning of sequential strategies.
\newblock \emph{CoRR}, abs/1905.03030, 2019.

\bibitem[Press et~al.(2022)Press, Smith, and Lewis]{press2022train}
Press, O., Smith, N.~A., and Lewis, M.
\newblock Train short, test long: Attention with linear biases enables input
  length extrapolation.
\newblock In \emph{{ICLR}}. OpenReview.net, 2022.

\bibitem[Reed et~al.(2022)Reed, Zolna, Parisotto, Colmenarejo, Novikov,
  Barth{-}Maron, Gimenez, Sulsky, Kay, Springenberg, Eccles, Bruce, Razavi,
  Edwards, Heess, Chen, Hadsell, Vinyals, Bordbar, and
  de~Freitas]{reed2022generalist}
Reed, S.~E., Zolna, K., Parisotto, E., Colmenarejo, S.~G., Novikov, A.,
  Barth{-}Maron, G., Gimenez, M., Sulsky, Y., Kay, J., Springenberg, J.~T.,
  Eccles, T., Bruce, J., Razavi, A., Edwards, A., Heess, N., Chen, Y., Hadsell,
  R., Vinyals, O., Bordbar, M., and de~Freitas, N.
\newblock A generalist agent.
\newblock \emph{CoRR}, abs/2205.06175, 2022.

\bibitem[Ritchie et~al.(2016)Ritchie, Horsfall, and Goodman]{ritchie2016deep}
Ritchie, D., Horsfall, P., and Goodman, N.~D.
\newblock Deep amortized inference for probabilistic programs.
\newblock \emph{CoRR}, abs/1610.05735, 2016.

\bibitem[Santoro et~al.(2016)Santoro, Bartunov, Botvinick, Wierstra, and
  Lillicrap]{santoro2016meta}
Santoro, A., Bartunov, S., Botvinick, M.~M., Wierstra, D., and Lillicrap, T.~P.
\newblock Meta-learning with memory-augmented neural networks.
\newblock In \emph{{ICML}}, volume~48 of \emph{{JMLR} Workshop and Conference
  Proceedings}, pp.\  1842--1850. JMLR.org, 2016.

\bibitem[Schmidhuber et~al.(1996)Schmidhuber, Zhao, and
  Wiering]{schmidhuber1996simple}
Schmidhuber, J., Zhao, J., and Wiering, M.
\newblock Simple principles of metalearning.
\newblock Technical report, IDSIA, 1996.

\bibitem[Thrun \& Pratt(1998)Thrun and Pratt]{thrun1998learning}
Thrun, S. and Pratt, L.~Y.
\newblock Learning to learn: Introduction and overview.
\newblock In \emph{Learning to Learn}, pp.\  3--17. Springer, 1998.

\bibitem[Vaswani et~al.(2017)Vaswani, Shazeer, Parmar, Uszkoreit, Jones, Gomez,
  Kaiser, and Polosukhin]{vaswani2017attention}
Vaswani, A., Shazeer, N., Parmar, N., Uszkoreit, J., Jones, L., Gomez, A.~N.,
  Kaiser, L., and Polosukhin, I.
\newblock Attention is all you need.
\newblock In \emph{{NIPS}}, pp.\  5998--6008, 2017.

\bibitem[Veness et~al.(2013)Veness, White, Bowling, and
  Gy{\"{o}}rgy]{veness2013partition}
Veness, J., White, M., Bowling, M., and Gy{\"{o}}rgy, A.
\newblock Partition tree weighting.
\newblock In \emph{{DCC}}, pp.\  321--330. {IEEE}, 2013.

\bibitem[Wang et~al.(2017)Wang, Kurth{-}Nelson, Soyer, Leibo, Tirumala, Munos,
  Blundell, Kumaran, and Botvinick]{wang2017learning}
Wang, J., Kurth{-}Nelson, Z., Soyer, H., Leibo, J.~Z., Tirumala, D., Munos, R.,
  Blundell, C., Kumaran, D., and Botvinick, M.~M.
\newblock Learning to reinforcement learn.
\newblock In \emph{CogSci}. cognitivesciencesociety.org, 2017.

\bibitem[Willems(1996)]{willems1996coding}
Willems, F. M.~J.
\newblock Coding for a binary independent piecewise-identically-distributed
  source.
\newblock \emph{{IEEE} Trans. Inf. Theory}, 42\penalty0 (6):\penalty0
  2210--2217, 1996.

\bibitem[Xie et~al.(2022)Xie, Raghunathan, Liang, and Ma]{xie2022explanation}
Xie, S.~M., Raghunathan, A., Liang, P., and Ma, T.
\newblock An explanation of in-context learning as implicit bayesian inference.
\newblock In \emph{{ICLR}}. OpenReview.net, 2022.

\bibitem[Zintgraf et~al.(2020)Zintgraf, Shiarlis, Igl, Schulze, Gal, Hofmann,
  and Whiteson]{zintgraf2020varibad}
Zintgraf, L.~M., Shiarlis, K., Igl, M., Schulze, S., Gal, Y., Hofmann, K., and
  Whiteson, S.
\newblock Varibad: {A} very good method for bayes-adaptive deep {RL} via
  meta-learning.
\newblock In \emph{{ICLR}}. OpenReview.net, 2020.

\end{thebibliography}
    \bibliographystyle{icml2023}

    \newpage
    \appendix
    \onecolumn
    
    \section{Architecture Ablation Study}
\label{sec:ablation-study}

We conducted an ablation study on the neural models we trained.
This was used both to select the best parameters for the main experiments, and better understand the impact of number of parameters and memory size on the models' capabilities. 
We trained all the networks on sequences of length~$256$, sampled from the \ptw{} prior.
We trained $5$ different seeds for each set of hyperparameters.
We ran each distribution-architecture-hyperparameter triplet on a single GPU on our internal cluster.

\paragraph{Vanilla RNNs and LSTMs}

For these networks, we swept over three hidden sizes: 64, 128 and 256.
We also swept over the number of dense layers to be appended after the recurrent core: 0, 1, or 2 layers.
These layers all contain 128 neurons.
The best performing models for both architectures were the largest ones, \ie 256 neurons in the recurrent core and 2 extra dense layers of 128 neurons after the core.
These are the hyperparameters we picked for the main study.
In \cref{fig:rnns-capacity-regret}, we plot the performance over the number of parameters of the model.
The performance is the averaged cumulative regret over 10k trajectories sampled from the \ptw{} prior.
The figure generally reveals a downward trend: the more parameters, the lower the prediction error.

\paragraph{Stack-RNNs/LSTMs}

For these networks, we performed the same sweep as for simple RNNs and LSTMs above.
In addition, we swept over the stack size (1, 8 or total sequence length, \eg 256) and the stack cell width (1, 2 and 8 dimensions).
The best performing models for both architectures were again the largest ones, \ie 256 neurons in the core and 2 extra dense layers of 128 neurons after the core.
These are the hyperparameters we picked for the main study.
\cref{fig:rnns-capacity-regret} also shows the same trend as for standard RNNs and LSTMs.
Furthermore, the best performing models for both architectures use a stack size of 8 and a stack cell size of 8 too.
This means that the networks cannot store the whole history of observations in the stack (at least not straightfowardly), but this size seems sufficient and smaller stacks might make training easier.

\paragraph{Transformers}

For these networks, we used an embedding size of 64 and 8 heads.
We swept over three positional encodings: classical sin/cos from the original Transformer paper, ALiBi which work well for short span dependencies, and the relative positional encodings from the TransformerXL paper.
We also swept over the number of layers: 2, 4, 8 and 16.
We first observe that all networks, regardless their size or positional encodings, train very well: The loss curves are smooth (not shown here) and the variance over seeds is small.
The best performing models are the ones using the relative positional encodings and the largest ones, \ie with 16 layers.
These are the hyperparameters we picked for the main study.
In \cref{fig:transformers-capacity-regret}, we report the performance over the capacity of the model, measured in number of parameters, for the different positional encodings.

\begin{figure}[htb]
    \centering
    \begin{minipage}{.48\textwidth}
        \centering
        \includesvg[scale=0.48]{figures/performance_parameters_rnns.svg}
        \caption{
            Cumulative regret (in bits) of the different RNNs, over their number of parameters.
            Dashed line shows \ptw$_8$.
        }
        \label{fig:rnns-capacity-regret}
    \end{minipage}
    \hspace{.02\textwidth}
    \begin{minipage}{.48\textwidth}
        \centering
        \includesvg[scale=0.48]{figures/performance_parameters_transformers.svg}
        \caption{
            Cumulative regret (in bits) of the different Transformers, over their number of parameters.
            Dashed line shows \ptw$_8$.
        }
        \label{fig:transformers-capacity-regret}
    \end{minipage}
\end{figure}

\clearpage

\section{Illustration of Data Generating Sources}
\label{sec:example-sequences}

\subsection{\ptw{} Switching-Point Statistics}
\label{sec:a_ptw_switchpoint_stats}

\paragraph{Positions of the Switching-Points}

\begin{figure}[htb]
    \centering
    \includesvg[scale=0.4]{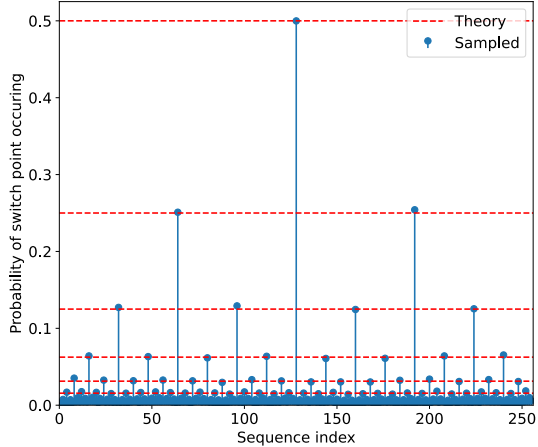}
    \caption{
        Distribution of \ptw{} switching-points, over sequence indexes, computed over 10000 sequences. 
        The length of the sequences is fixed to 256.
    }
    \label{fig:positions-switches}
\end{figure}

To give a better intuition on where the \ptw{} switching-points occur in the sequence, we plot their distribution in \cref{fig:positions-switches}.
They are mostly present at half the sequence (probability 1/2), then at all the quarters of the sequence (probability 1/4), and so on, dividing by two the intervals recursively (and dividing the probability by 2).

\paragraph{Number of switching-points}

We are also interested in the prior distribution of the number of switching-points, which is implicitly given by the \ptw{} prior.
We can get recursive and explicit formulas as follows:
The recursive definition of the \ptw{} distribution \cite{veness2013partition}
\begin{align}\label{eq:ptwrec}
  {\normalfont\textsc{ptw}}_d(x_{1:n}) = \textstyle{\frac12}\rho(x_{1:N}) + \textstyle{\frac12}{\normalfont\textsc{ptw}}_{d-1}\rho(x_{1:2^{d-1}}){\normalfont\textsc{ptw}}_{d-1}\rho(x_{2^{d-1}+1:n})
\end{align}
leads to the following recursion for the probability of $k$ switching-points for $d\geq 1$
\begin{align}\label{eq:Pdkrec}
  P_d[k] ~=~ \frac12\delta_{k,0} + \sum_{l=1}^{k-1} P_{d-1}[k-l]\cdot P_{d-1}[l-1], ~~~~~\text{and}~~~~~ P_0[k]=\delta_{k,0}
\end{align}
(the number of switching-points is the number in the left half plus the number in the right half plus 1).
From this we can compute $P_d[k]$ in time $O(d\cdot k_{\max})$.
We plot the curves for $d=0,\ldots,9,\infty$ in \cref{fig:theoretical-switches}.
We also plot the same curves in \cref{fig:empirical-switches}, but from empirically sampling from our \ptw{} data source and counting the number of switching-points.
We sample 10 batches of 1000 sequences and report the mean and standard deviations of number of switching-points.
The match is very good, with a very little statistical error.

\begin{figure}[htb]
    \centering
    \begin{minipage}{.48\textwidth}
        \centering
        \includesvg[scale=0.48]{figures/ptw_switches_theory.svg}
        \caption{
            Theoretical \ptw{} distribution of number of switches. 
            For $k\geq d=0,\ldots,9$ (colored curves).
            For $k<d$, $P_d[k]=P_\infty[k]$ (black curve).
        }
        \label{fig:theoretical-switches}
    \end{minipage}
    \hspace{.02\textwidth}
    \begin{minipage}{.48\textwidth}
    \centering
    \includesvg[scale=0.48]{figures/ptw_switches_sampled.svg}
    \caption{
        \ptw{} empirical distribution of number of switches over 10 batches of 1000 sequences each (colored curves). 
        We also added the theoretical case $P_\infty[k]$ (black curve).
    }
    \label{fig:empirical-switches}
    \end{minipage}
\end{figure}

The empirically observed kink at $k=d$ is indeed real for small $d$ and gets washed out for larger $d$.
It is easy to see from the recursion and from the plot that $P_d[k]$ is the same for all $d>k$.
We can hence compute the limit for $d\to\infty$: A sequence with $k$ switches corresponds to a full binary tree with $k$ inner-switch nodes and $k+1$ leaves-segments.
\ptw{} assigns a probability $1/2$ to each decision of whether to switch or not.
Therefore for such a partition $\cal P$ we have $2^{-\Gamma_d(\cP)}=\smash{(\frac12)^{k+(k+1)}}$.
There are $C(k)$ such trees, where $C(k)=\smash{\frac{(2k)!}{k!(k+1)!}}=[1, 1, 2, 5, 14, 42, 132, \ldots]$ are the Catalan numbers. 
Therefore
\begin{align}
  P_\infty[k] ~=~ C(k)\cdot 2^{-\Gamma_d({\cal P})} ~=~ \textstyle{\frac{(2k)!}{k!(k+1)!}2^{-2k-1}} ~=~ P_d[k] ~~~\text{for}~~~ d>k
\end{align}
This expression can also be verified by inserting it into \eqref{eq:Pdkrec}, using binomial identities.
For large $k$, Stirling approximation gives $P_\infty[k]\approx k^{-3/2}/2\sqrt\pi$, which is quite accurate even for $k$ as low as $1$.
This is good news: The prior distribution of switches is as close to non-dogmatic as possible: $1/k$ would not sum, $1/k^2$ is quite good, $1/k^{1.5}$ is even better, while $1/2^k$ would be very dogmatic and therefore bad.
This good behavior is not a priori obvious. 
Indeed, if in \ptw{} we would choose the switch probability $p$ anything but $1/2$ (larger or smaller!),  $P_{\infty,p}[k]=P_\infty[k]\cdot 2(1-p)\cdot[4p(1-p)]^k$ which decreases exponentially in $k$ for $p\neq 1/2$.
From \eqref{eq:ptwrec}, we can also derive the expected number of switching-points
\begin{align}\label{eq:expected_num_switches}
  \mathbb{E}_d[k] ~=~ \textstyle\frac12\cdot 0 + \frac12(1+\mathbb{E}_{d-1}[k]+\mathbb{E}_{d-1}[k]) ~=~ \frac12+\mathbb{E}_{d-1}[k]~=\ldots=~d/2  
\end{align}
which grows linearly with $d$ (as expected) due to the tail of $P_d[k]$ being dragged out for $d\to\infty$.
Similarly for $p\neq 1/2$ we have 
\begin{align*}
  \mathbb{E}_d[k] ~&=~ (1-p)\cdot 0 + p(1+\mathbb{E}_{d-1}[k]+\mathbb{E}_{d-1}[k])
  ~=~ p+2p\cdot\mathbb{E}_{d-1}[k]~=\ldots \\
  ~&~~~~~~~\ldots=~ p\cdot[1+2p+(2p)^2+\ldots+(2p)^{d-1}] ~=~ p\frac{1-(2p)^d}{1-2p}
  ~~~\stackrel{d\to\infty}\longrightarrow~~~ \left\{\genfrac{}{}{0pt}{} {\frac{p}{1-2p}~\text{for}~p<\frac{1}{2}}{\frac{p}{2p-1}(2p)^d~\text{for}~p>\frac{1}{2}} \right. 
\end{align*}
That is, for $p<\frac{1}{2}$ this implies a prior believe of $k$ (strongly) peaked around $\frac{p}{1-2p}$, not growing with $d$,
while for $p>\frac{1}{2}$, it increases exponentially in $d$: $k\propto (2p)^d=n^\alpha$ with $0<\alpha:=\log_2(2p)<1$.

\subsection{Switching-Point Statistics for Other Priors}
\label{sec:a_other_switchpoint_stats}

An example draw from the \lin{} prior is shown in \cref{fig:single_traj_lin}.
Empirical switching-point statistics are in \cref{fig:hist-no-sps-lin} and \cref{fig:hist-loc-sps-lin}.

An example draw from the Random Uniform prior is shown in \cref{fig:single_traj_random}.
Empirical switching-point statistics are in \cref{fig:hist-no-sps-random} and \cref{fig:hist-loc-sps-random}.

An example draw from the Random Periodic prior is shown in \cref{fig:single_traj_regular}.
Empirical switching-point statistics are in \cref{fig:hist-no-sps-regular} and \cref{fig:hist-loc-sps-regular}.

\begin{figure}[htb]
    \centering
    \begin{minipage}[b]{\textwidth}
      \centering
      \includesvg[scale=0.44]{figures/single_traj_lin.svg}
      \caption{Example draw from \lin{} prior and model predictions.}
      \label{fig:single_traj_lin}
    \end{minipage}
    \begin{minipage}[b]{\textwidth}
      \centering
      \includesvg[scale=0.44]{figures/single_traj_random.svg}
      \caption{Example draw from Random Uniform prior ($\text{Uniform}(1, 256)$) and model predictions.}
      \label{fig:single_traj_random}
    \end{minipage}
    \begin{minipage}[b]{\textwidth}
      \centering
      \includesvg[scale=0.44]{figures/single_traj_regular.svg}
      \caption{Example draw from Random periodic prior (period=$20$ steps) and model predictions.}
      \label{fig:single_traj_regular}
    \end{minipage}
\end{figure}

\begin{figure}[htb]
    \centering
    \begin{minipage}[b]{.48\textwidth}
      \centering
      \includesvg[scale=0.4]{figures/hist_no_switchpoints_lin.svg}
      \caption{No. of switching-points per sequence (\lin{} prior).}
      \label{fig:hist-no-sps-lin}
    \end{minipage}%
    \hspace{.02\textwidth}
    \begin{minipage}[b]{.48\textwidth}
      \centering
      \includesvg[scale=0.4]{figures/hist_loc_switchpoints_lin.svg}
      \caption{Switching-point locations (\lin{} prior).}
      \label{fig:hist-loc-sps-lin}
    \end{minipage}
    \begin{minipage}[b]{.48\textwidth}
      \centering
      \includesvg[scale=0.4]{figures/hist_no_switchpoints_random.svg}
      \caption{No. of switching-points per sequence (Random Uniform prior, $\text{Uniform}(1, 256)$).}
      \label{fig:hist-no-sps-random}
    \end{minipage}%
    \hspace{.02\textwidth}
    \begin{minipage}[b]{.48\textwidth}
      \centering
      \includesvg[scale=0.4]{figures/hist_loc_switchpoints_random.svg}
      \caption{Switching-point locations (Random Uniform prior, $\text{Uniform}(1, 256)$).}
      \label{fig:hist-loc-sps-random}
    \end{minipage}
    \begin{minipage}[b]{.48\textwidth}
      \centering
      \includesvg[scale=0.4]{figures/hist_no_switchpoints_regular.svg}
      \caption{No. of switching-points per sequence (Regular Periodic prior, period=$20$ steps).}
      \label{fig:hist-no-sps-regular}
    \end{minipage}%
    \hspace{.02\textwidth}
    \begin{minipage}[b]{.48\textwidth}
      \centering
      \includesvg[scale=0.4]{figures/hist_loc_switchpoints_regular.svg}
      \caption{Switching-point locations (Regular Periodic prior, period=$20$ steps).}
      \label{fig:hist-loc-sps-regular}
    \end{minipage}
\end{figure}

\subsection{Models' Regret Along the Sequences}
\label{sec:regret_along_sequences}

\begin{figure}[htb]
    \centering
    \includesvg[scale=0.6]{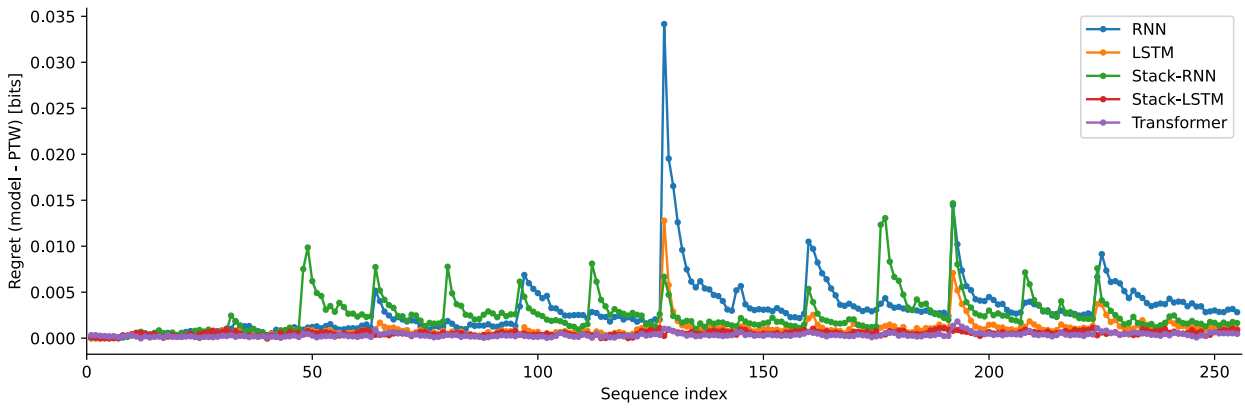}
    \caption{Difference of the average regret per sequence index, over 10000 sequences of length 256, drawn from the \ptw{} prior.}
    \label{fig:mean-sequence-index-diff}
\end{figure}

In \cref{fig:mean-sequence-index} we plot the average regret of the different models for all sequence indexes on 10000 sequences of length 256, drawn from the \ptw{} prior.
The models have also been trained on this prior.
The match is almost perfect.
We also plot the difference between the models' regret and \ptw's regret in \cref{fig:mean-sequence-index-diff}, to emphasize the models' relative performance.
Note that in theory, the models can do better than \ptw{} on some indexes, but not when summing over all of them.

\begin{figure}[htb]
    \centering
    \includesvg[scale=0.6]{figures/mean_sequence_index.svg}
    \caption{Average regret per sequence index, over 10000 sequences of length 256, drawn from the \ptw{} prior.}
    \label{fig:mean-sequence-index}
\end{figure}

\clearpage

\section{Additional Experiments}
\label{sec:additional-experiments}

\subsection{On-Distribution Performance}
\label{sec:off-distribution-random}

\begin{figure}[htb]
    \centering
    \includesvg[scale=0.55]{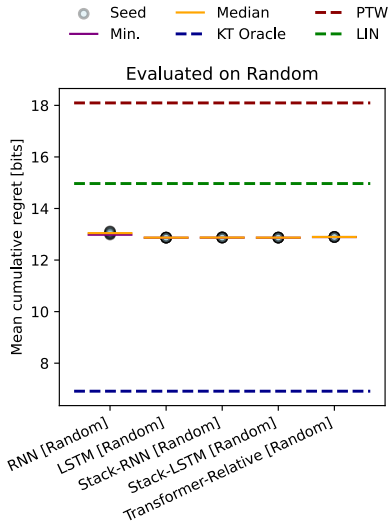}
    \caption{
        On-distribution evaluation ($10$k sequences, length~$256$). 
        Models were trained and evaluated on data from the Random Uniform distribution ($\text{Uniform}(1, 256)$) over segment lengths. 
        Note that we have no known exact Bayesian inference baseline in this case, though \lin{} comes with certain robustness guarantees that ensure good prediction performance in this setting.
        Neural networks trained precisely on this data distribution manage to outperform \lin{} though.
    }
    \label{fig:main_result_random}
\end{figure}

\cref{fig:main_result_random} shows the models' performance for training and evaluating on data with segment lengths drawn from a Random Uniform prior.

\subsection{Off-Distribution Evaluation}
\label{sec:off-distribution-generalization}

\cref{fig:generalisation_regular} shows how models trained on data from the \ptw{} and \lin{} priors generalize to evaluating on data that follows Regular Periodic shifts.
\cref{fig:generalisation_random_ptw} and \cref{fig:generalisation_random_lin} show how models trained on Random Uniform segment lengths behave when evaluated on data from the \ptw{} and \lin{} priors, respectively.

\begin{figure}[htb]
    \begin{minipage}[t]{.38\textwidth}
        \centering
        \includesvg[scale=0.45]{figures/generalization_ptw_lin_on_regular.svg}
        \caption{
            Off-distribution evaluation ($10$k sequences, length~$256$).
            The models' training distribution indicated in the square brackets.
            All models are evaluated with regular periodic segment lengths of period~$20$.
            Red dashed line shows \ptw$_8$.
        }
        \label{fig:generalisation_regular}
    \end{minipage}
    \hspace{.02\textwidth}
    \begin{minipage}[t]{.28\textwidth}
        \centering
        \includesvg[scale=0.45]{figures/generalization_random_on_ptw.svg}
        \caption{
            Off-distribution evaluation ($10$k sequences, length~$256$).
            Models were trained on data from Random Uniform segment lengths ($\text{Uniform}(1, 256)$) and evaluated on data from \ptw$_8$.
        }
        \label{fig:generalisation_random_ptw}
    \end{minipage}
    \hspace{.02\textwidth}
    \begin{minipage}[t]{.28\textwidth}
        \centering
        \includesvg[scale=0.45]{figures/generalization_random_on_lin.svg}
        \caption{
            Off-distribution evaluation ($10$k sequences, length~$256$).
            Models were trained on data from Random Uniform segment lengths ($\text{Uniform}(1, 256)$) and evaluated on data from \lin.
        }
        \label{fig:generalisation_random_lin}
    \end{minipage}
\end{figure}

\subsection{Evaluation on Longer Sequence Lengths at Test Time}
\label{sec:longer-lenght-eval}

\begin{figure}[htb]
    \centering
    \includesvg[scale=0.6]{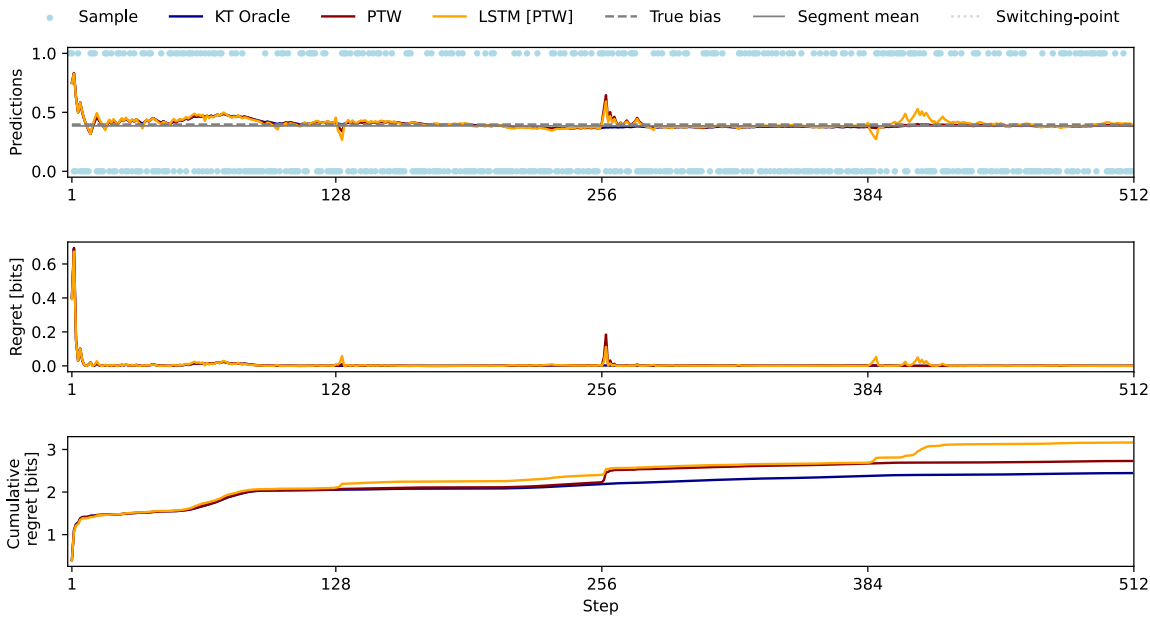}
    \caption{
        Sequence-length generalization: single sequence of length~$512$ without switching points (which is quite likely under \ptw$_9$ prior).
        The LSTM predictions shown are taken from a model trained on sequences of length~$256$ (from \ptw$_8$ prior).
        The LSTM generalizes well to sequences of longer length, taking the main hit in terms of cumulative regret (compared to \ptw) around step~$128$, which is the most likely switching-point on the data that the model was trained on, and step~$384$ (which is a multiple of~$128$).
        Otherwise, predictions remain stable despite the sequence being twice as long as any sequence the model has ever experienced during training (which is an indicator that internal dynamics remain stable too).
    }
    \label{fig:single_traj_generalisation_ptw_lstm_512}
\end{figure}

\begin{figure}[htb]
    \centering
    \includesvg[scale=0.6]{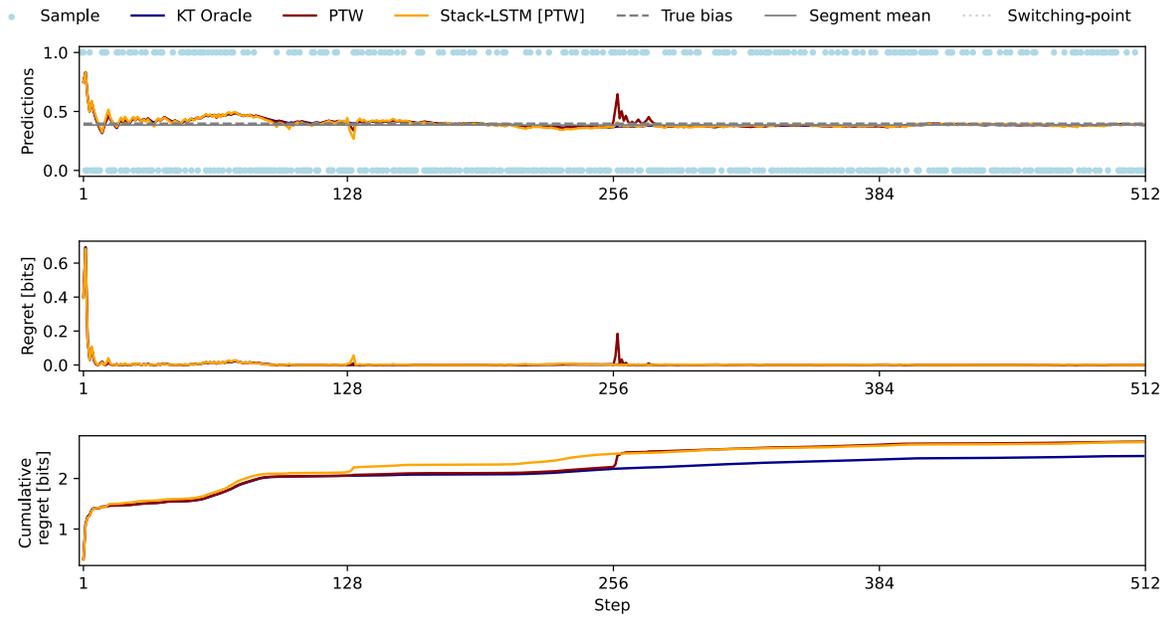}
    \caption{
        Same as \cref{fig:single_traj_generalisation_ptw_lstm_512} but model shown here is Stack-LSTM.Compared to the plain LSTM, the Stack-LSTM seems to predict a change point at step $384$ with lower probability. 
    }
    \label{fig:single_traj_generalisation_ptw_stack_lstm_512}
\end{figure}

\begin{figure}[htb]
    \centering
    \includesvg[scale=0.6]{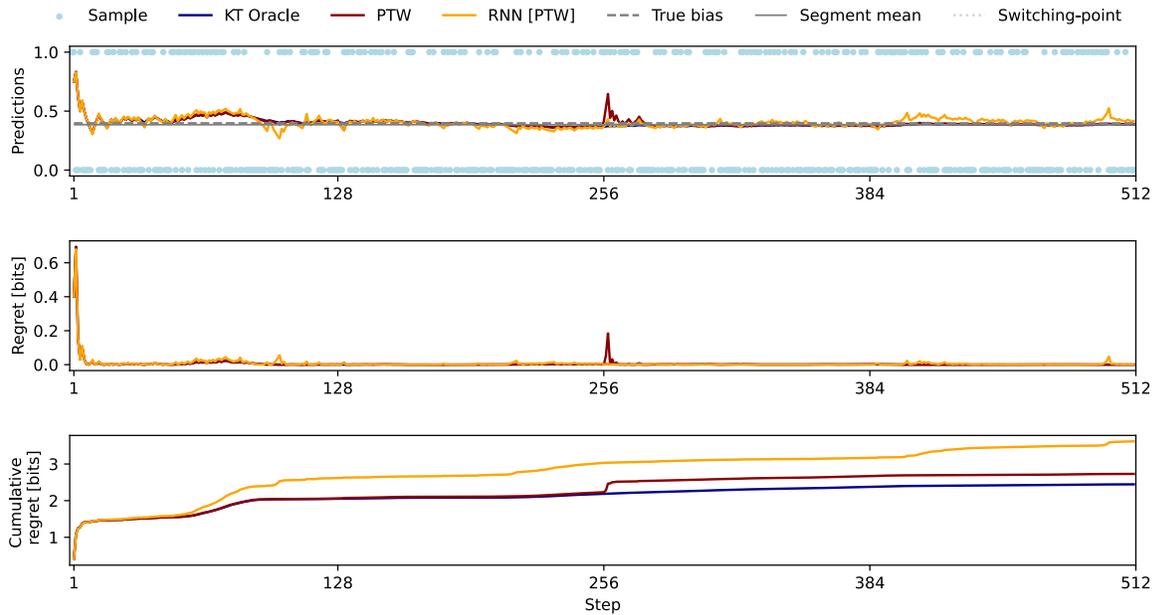}
    \caption{
        Same as \cref{fig:single_traj_generalisation_ptw_lstm_512} but model shown here is RNN. Compared to the LSTM, the RNN predictions are a bit worse on this sequence, but internal dynamics seem to remain very stable far beyond the training range of $256$ steps.   
    }
    \label{fig:single_traj_generalisation_ptw_rnn_512}
\end{figure}

\begin{figure}[htb]
    \centering
    \includesvg[scale=0.6]{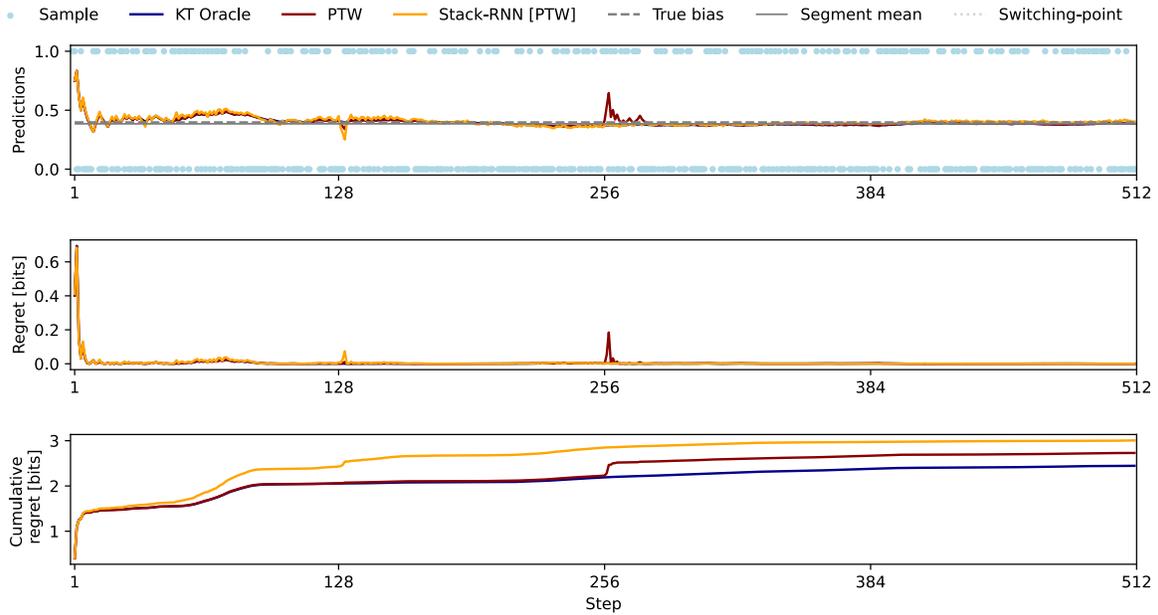}
    \caption{
        Same as \cref{fig:single_traj_generalisation_ptw_rnn_512} but model shown here is Stack-RNN. It is hard to identify a qualitative difference to the plain RNN; the Stack-RNN performs better / more stable in the second half of the trajectory, which is in line with the trend seen for the Stack-LSTM in \cref{fig:single_traj_generalisation_ptw_stack_lstm_512} compared to the plain LSTM.
    }
    \label{fig:single_traj_generalisation_ptw_stack_rnn_512}
\end{figure}

\begin{figure}[htb]
    \centering
    \includesvg[scale=0.6]{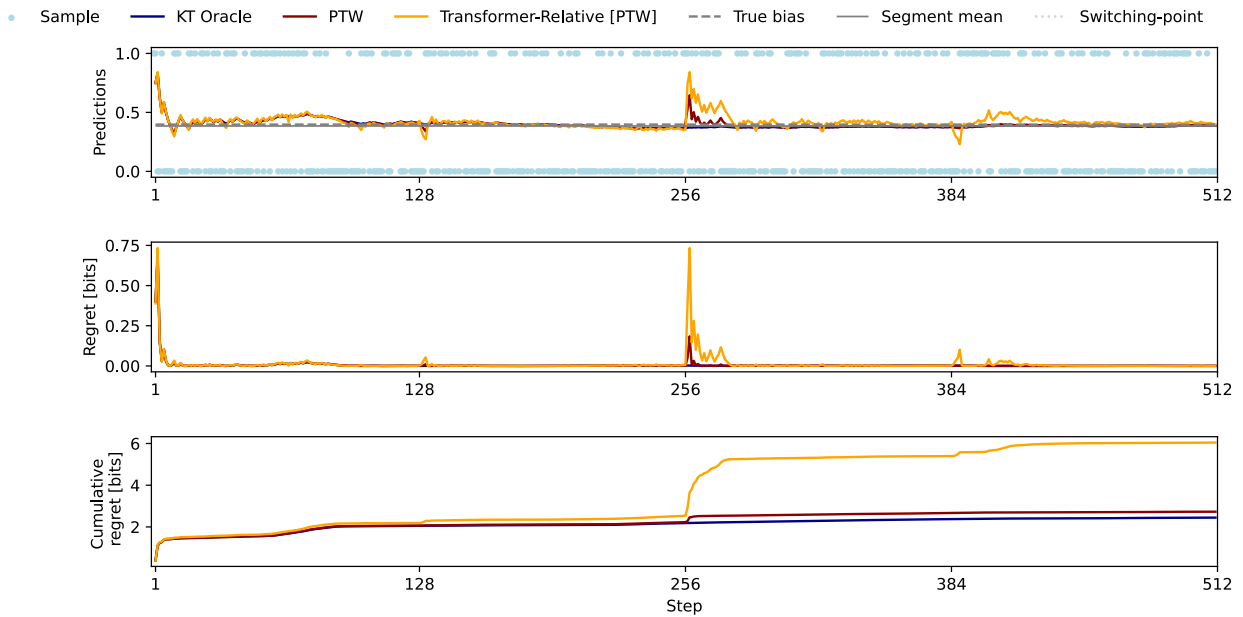}
    \caption{
        Same as \cref{fig:single_traj_generalisation_ptw_lstm_512} but model shown here is Transformer-Relative. Compared to all other neural models, the transformer seems to struggle with predicting well from step~$256$ onward (note that the model was trained with sequences of length~$256$). 
    }
    \label{fig:single_traj_generalisation_ptw_transformer_relative_512}
\end{figure}

\begin{figure}[htb]
    \centering
    \includesvg[scale=0.5]{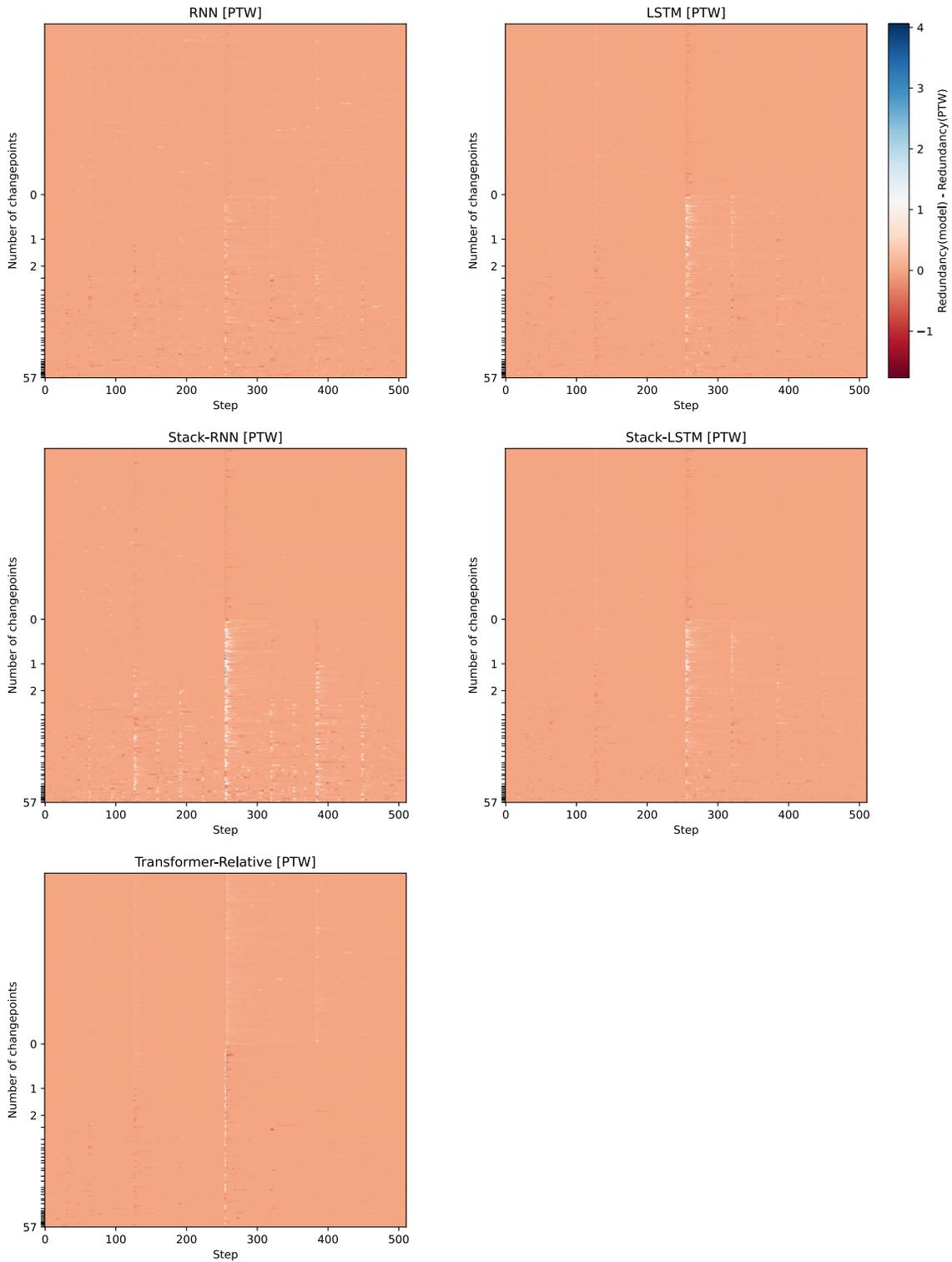}
    \caption{
        Models evaluated on $500$ trajectories of length~$512$ drawn from \ptw{}$_9$ prior. Models trained on sequences of length~$256$ drawn from \ptw{}$_8$. In each panel: each row is a single trajectory, and the color encodes the difference in redundancy between the model minus \ptw{}$_9$. Trajectories are ordered by the number of switching-points (y-axis). See main text for a discussion of the figure.
    }
    \label{fig:single_traj_generalisation_analysis_512}
\end{figure}

See \cref{fig:single_traj_generalisation_ptw_lstm_512},  \cref{fig:single_traj_generalisation_ptw_stack_lstm_512}, \cref{fig:single_traj_generalisation_ptw_rnn_512}, \cref{fig:single_traj_generalisation_ptw_stack_rnn_512}, and \cref{fig:single_traj_generalisation_ptw_transformer_relative_512}  for example sequences of length generalization of the different models.
For a large-scale quantitative evaluation see \cref{fig:generalisation-longer-lengths} in the main text.
Finally, \cref{fig:single_traj_generalisation_analysis_512} gives some insight into generalization behavior of the different models. In the figure, models were trained on sequences of length~$256$ drawn from \ptw{}$_8$, but evaluated on sequences of length~$512$ drawn from \ptw{}$_9$.
In that case, the most likely change point occurs at~$256$, but since models were trained on trajectories of length~$256$ all models, except the transformer predict better than \ptw{}$_9$ if no change point occurs (for all trajectories with 0 switching-points, roughly the upper half of each panel, there is a dark red band at~$256$). If the most likely change point actually occurs (trajectories with $1$ or more switching-points), neural models predict the change at~$256$ with lower probability than \ptw{}$_9$, leading to a white/blue band in the lower half of each panel. Similar trends are also seen for other highly likely switching-points such as $128$ or $384$, with the Stack-RNN showing the strongest white bands (consistent with having the worst performance in \cref{fig:generalisation-longer-lengths}).

\clearpage

\section{Proof of \cref{thm:limitations}}
\label{subsec:proof-4.2}

\begin{proof}
    By way of contradiction, assume $\mathbb{E}_{\mu}|\nu_\Theta(a_t|x_{<t})-\xi(a_t|x_{<t})|\to 0~\forall\mu\in\cM$. 
    In particular this implies
    \begin{align*}
      \mathbb{E}_{\mu_i}|\nu_{\theta_t}(a_t)- \xi(a_t|x_{<t})| ~\longrightarrow~ 0
    \end{align*}
    where we have used $\nu_\Theta(a_t|x_{<t})=\nu_{\theta_t}(a_t)$.
    Combining this with Solomonoff's theorem \cite{hutter2005universal} (Thm.3.19iii)
    \begin{align*}
      \mathbb{E}_{\mu_i}|\xi(a_t|x_{<t})-\mu_i(a_t|x_{<t})| ~\longrightarrow~ 0
    \end{align*}
    we get
    \begin{align*}
      |\nu_{\theta_t}(a_t)-\mathbb{E}_{\mu_i}\mu_i(a_t|x_{<t})|
      ~\leq~ \mathbb{E}_{\mu_i}|\nu_{\theta_t}(a_t)-\mu_i(a_t|x_{<t})| ~\longrightarrow~ 0
    \end{align*}
    The inequality exploits that $\nu_\Theta$ is memoryless. 
    Finally combining these convergences for $i=1$ and $i=2$ we get
    \begin{align*}
      |\mathbb{E}_{\mu_1}\mu_1(a_t|x_{<t})-\mathbb{E}_{\mu_2}\mu_2(a_t|x_{<t})| ~\longrightarrow~ 0
    \end{align*}
    which contradicts the theorem's assumption on $\mu_i$.
\end{proof}


\section{Prior Sampling Algorithms}
\label{sec:prior_sampling}

This section provides more detail on how the temporal partitions are sampled under the \ptw{} and \lin{} priors which are defined in \cref{sec:priors_and_baselines}.
Both priors are hierarchical in the sense that they first define a prior on the latent switching-point structure, and then assign a $Beta(0.5, 0.5)$ prior to the Bernoulli process governing each segment.
Here we focus just on the non-trivial first stage of each hierarchical process.

\subsection{Sampling From the \ptw{} Prior}

Given a fixed $d$, \cref{alg:ptw_prior_sample} samples a binary temporal partition from $\cC_d$ distributed according to the \ptw{} prior when invoked with an offset $o=0$.
The algorithm works by first flipping a fair coin which determines whether or not to continue splitting the current segment in half; in the case of a split, the process continues recursively on the two half segments.
The base case is handled by $d=0$ which corresponds to a segment consisting of a single time point, which obviously cannot be split further.
The expected running time is proportional to the expected number of switches, which we show in \cref{sec:a_ptw_switchpoint_stats}, \cref{eq:expected_num_switches} to be equal to $\tfrac{d}{2} = O(\log n)$.

\begin{algorithm}[h!]
    \caption{$\text{\sc tps}_d(o)$}
    \label{alg:ptw_prior_sample}
    \begin{algorithmic}
        \REQUIRE An offset $o \in \mathbb{N}$
        \IF{$d = 0$}
        \STATE \textbf{return} $\{(o+1, o+1) \}$
        \ENDIF
        \STATE Sample $r \sim {\sc Bernoulli}(0.5)$
        \IF{$r = 0$}
        \STATE \textbf{return} $\{(o + 1, o+2^d) \}$
        \ELSE
        \STATE \textbf{return} $\text{\sc tps}_{d-1}(o) \cup \text{\sc tps}_{d-1}(o + 2^{d-1})$
        \ENDIF
    \end{algorithmic}
\end{algorithm}

\subsection{Sampling From the \lin{} Prior}

\cref{alg:lin_prior_sample} samples a temporal partition from $\cP_n$ distributed according to the \lin{} prior.
The algorithm starts in state $(1,1)$, with the left component representing the current time, and the right component representing the time index of the current segment.
The current state $(t,t_c)$ is adapted $n$ times, where $\frac{1/2}{t - t_c +1}$ gives the probability of a change-point occurring at time $t$.
The worst-case runtime complexity of this algorithm is clearly linear in $n$.

\begin{algorithm}[h!]
    \caption{$\text{\sc lin-prior-sample}(n)$}
    \label{alg:lin_prior_sample}
    \begin{algorithmic}
        \REQUIRE Sequence length $n \in \mathbb{N}$
        \STATE $t \leftarrow 1$, $t_c \leftarrow 1$, $\mathcal{T} \leftarrow \{ \}$ 
        \WHILE{$t < n$}
            \STATE Sample $r \sim \text{\sc Bernoulli}\left(\frac{1/2}{t - t_c +1}\right)$
            \IF{$r = 1$}
                \STATE $\mathcal{T} \leftarrow \mathcal{T} \cup \{ (t_c, t) \}$
                \STATE $t_c = t+1$
            \ENDIF
            \STATE $t \leftarrow t+1$
        \ENDWHILE
        \STATE $\mathcal{T} \leftarrow \mathcal{T} \cup \{ (t_c, t) \}$
        \STATE \textbf{return} $\mathcal{T}$
    \end{algorithmic}
\end{algorithm}


\section{Discrete Bayesian Mixtures}
\label{sec:a_bayesmix}

A fundamental technique for constructing algorithms that work well under the logarithmic loss is Bayesian model averaging.
Given a non-empty discrete set of probabilistic data generating sources $\cM := \{\rho_1, \rho_2, \dots \}$ and a prior weight $w^\rho_0 > 0$ for each $\rho \in \cM$ such that $\sum_{\rho \in \cM} w^\rho_0 = 1$, the Bayesian mixture predictor is defined in terms of its marginal by $\xi(x_{1:n}) := \sum_{\rho \in \cM} w^\rho_0 \, \rho(x_{1:n})$.
The predictive probability is thus given by the ratio of the marginals $\xi(x_n | x_{<n}) = \xi(x_{1:n}) / \xi(x_{<n})$.
The predictive probability can also be expressed in terms of a convex combination of conditional model predictions, with each model weighted by its posterior probability.
More explicitly,
\begin{align*}
    \xi(x_n \cdbar x_{<n}) &=
    \frac{\sum_{\rho \in \cM} w^\rho_0 \, \rho(x_{1:n})}{\sum_{\rho \in \cM} w^\rho_0 \, \rho(x_{<n})}\\
    &= \sum_{\rho \in \cM} w^\rho_{n-1} \, \rho(x_n \cdbar x_{<n}) \\
    \text{where}~~~~~ w^\rho_{n-1} &:= \frac{w^\rho_0 \, \rho(x_{<n})}{\sum_{\nu \in \cM} w^\nu_{0} \, \nu(x_{<n})}.
\end{align*}

A fundamental property of Bayesian mixtures is that if there exists a model $\rho^{*} \in \cM$ that predicts well, then $\xi$ will predict well since the cumulative loss satisfies
\begin{align}\label{eq:bayes_mixture_regret}
    - \log \xi(x_{1:n}) &= - \log \sum_{\rho \in \cM} w^\rho_0 \, \rho(x_{1:n})\notag\\ &\leq -\log  w^{\rho^*}_0  \rho^*(x_{1:n})\notag\\
    &= \log \left(\tfrac{1}{w^{\rho^*}_0} \right) -\log\rho^*(x_{1:n}).
\end{align}
\cref{eq:bayes_mixture_regret} implies that a constant regret bounded by $\log (1/w^{\rho^*}_0)$ is suffered when using $\xi$ in place of the best (in hindsight) model $\rho^*\in\cM$.

\end{document}